\documentclass[lettersize,journal]{IEEEtran}
\usepackage[none]{hyphenat}
\usepackage{amsmath,amsfonts}
\usepackage{algorithmic}
\usepackage{array}
\usepackage[caption=false,font=normalsize,labelfont=sf,textfont=sf]{subfig}
\usepackage{textcomp}
\usepackage{stfloats}
\usepackage{url}
\usepackage{verbatim}
\usepackage{graphicx}
\usepackage{booktabs}   % 解决 \toprule, \midrule, \bottomrule 报错
\usepackage{graphicx}   % 解决 \resizebox 报错
\usepackage{multirow}   % 解决多行合并报错
\usepackage{hyperref}
\hypersetup{colorlinks=true, linkcolor=blue}

\def\BibTeX{{\rm B\kern-.05em{\sc i\kern-.025em b}\kern-.08em
    T\kern-.1667em\lower.7ex\hbox{E}\kern-.125emX}}
\usepackage{balance}
\begin{document}
\title{Latent World Models for Automated Driving: A Unified Taxonomy, Evaluation Framework, and Open Challenges}

\author{%
Rongxiang~Zeng and Yongqi~Dong$^{*}$%
%Rongxiang~Zeng$^{*}$ and Yongqi~Dong$^{\dagger}$%

%\thanks{$^{*}$First Author, $^{\dagger}$Corresponding Author.} Manuscript submitted 28 February 2026.

\thanks{ ($^{*}$\textit{Corresponding author: Yongqi Dong.}) 

Rongxinag Zeng is with RWTH Aachen University, Templergraben 55, 52062 Aachen, Germany (e-mail: \href{mailto: rongxiang.zeng@rwth-aachen.de}{rongxiang.zeng@rwth-aachen.de}).

Yongqi Dong is with Delft University of Technology, 2628 CN Delft, The Netherlands (e-mail: \href{mailto: yongqi.dong@rwth-aachen.de}{yongqi.dong@rwth-aachen.de})

Digital Object Identifier 10.1109/TITS.20XX.XXXXXXX}%
}

\markboth{IEEE TRANSACTIONS ON INTELLIGENT TRANSPORTATION SYSTEMS, VOL. XX, NO. XX}%
{Latent World Models for Automated Driving: A Unified Taxonomy, Evaluation Framework, and Open Challenges}

\maketitle

\begin{abstract}
\sloppy{}Emerging generative world models and vision-language-action (VLA) systems are rapidly reshaping automated driving by enabling scalable simulation, long-horizon forecasting, and capability-rich decision making. Across these directions, latent representations serve as the central computational substrate: they compress high-dimensional multi-sensor observations, enable temporally coherent rollouts, and provide interfaces for planning, reasoning, and controllable generation. This paper proposes a unifying latent-space framework that synthesizes recent progress in world models for automated driving. The framework organizes the design space by the target and form of latent representations (latent worlds, latent actions, latent generators; continuous states, discrete tokens, and hybrids) and by structural priors for geometry, topology, and semantics. Building on this taxonomy, the paper articulates five cross-cutting internal mechanics (i.e, structural isomorphism, long-horizon temporal stability, semantic and reasoning alignment, value-aligned objectives and post-training, as well as adaptive computation and deliberation) and connects these design choices to robustness, generalization, and deployability. The work also proposes concrete evaluation prescriptions, including a closed-loop metric suite and a resource-aware deliberation cost, designed to reduce the open-loop / closed-loop mismatch. Finally, the paper identifies actionable research directions toward advancing latent world model for decision-ready, verifiable, and resource-efficient automated driving.

\end{abstract}

\begin{IEEEkeywords}
World models, latent representations, automated driving, generative modeling, closed-loop evaluation.
\end{IEEEkeywords}

\section{Introduction}
\sloppy{}
\IEEEPARstart{A}UTOMATED driving couples high-dimensional multi-sensor perception with long-horizon decision making under stringent safety constraints. While large-scale real-world driving logs have accelerated learning-based autonomy \cite{wang_drivedreamer_2025, tang_omnigen_2025}, safety-critical interactions such as rare events and adversarial maneuvers remain sparse and expensive to validate in closed loop \cite{peng_safety-critical_2025, li_think2drive_2024, yang2025resim}, and purely synthetic simulators introduce nontrivial sim-to-real gaps \cite{wang_drivedreamer_2025, yang2025resim, 10242366}. Moreover, human-centered and socially compliant behavior adds further structural requirements beyond physical safety, demanding models that account for interaction norms and cooperative intent \cite{10394462, pnas1820676116, DONG2025100207}. These realities motivate representations that compress raw observations, support temporally coherent ``imagination" and prediction, and expose decision-relevant structure for downstream planning and control.

% Early learning-based pipelines demonstrated that perception can be mapped directly to control, and later imitation and hybrid approaches improved robustness through richer supervision and targeted scenario synthesis; nevertheless, they also revealed brittleness under distribution shift and long-tail interactions~\cite{codevilla2017cil, bansal2018chauffeurnet, chen2024end, arasteh2025validity, hallgarten2024can, ge2025unraveling}. 
% In parallel, classical reinforcement learning (RL) formalized control as planning over learned system dynamics, typically operating on explicitly modeled state transitions and reward structures \cite{10.1561/2200000086, 10422488, 10422159, Yan2025ADR1CR}. More recently, research on learned dynamics and latent world models extended this paradigm by learning compact, high-dimensional latent state representations directly from raw sensory inputs \cite{ha2018worldmodels,hafner2024masteringdiversedomainsworld,gumbsch2023learning,saanum2024simplifying}. Instead of planning in high-dimensional observation space, these approaches perform rollouts within learned latent dynamics that support spatiotemporal coherence and long-horizon reasoning, indicating that the structure and evolution of intermediate latent representations are often the dominant factors governing stability, controllability, and generalization.

Early learning-based pipelines demonstrated direct perception-to-control mapping, and later imitation and hybrid approaches improved robustness through richer supervision and targeted scenario synthesis; however, they also exposed brittleness under distribution shift and long-tail interactions~\cite{codevilla2017cil,bansal2018chauffeurnet,chen2024end,arasteh2025validity,hallgarten2024can,ge2025unraveling}. 
In parallel, reinforcement learning (RL) frames driving as sequential decision making with explicit transition and reward structure \cite{10.1561/2200000086,10422488,10422159,Yan2025ADR1CR}. Learned dynamics and latent world models extend this view by inferring compact latent states from raw sensory inputs and enabling imagination-based rollouts in latent space~\cite{ha2018worldmodels,hafner2024masteringdiversedomainsworld,gumbsch2023learning,saanum2024simplifying}. This shift suggests that the structure and temporal evolution of intermediate latent representations are often decisive for stability, controllability, and generalization.

Latent spaces have therefore emerged as a unifying computational substrate. They act as compact information bottlenecks that compress high-dimensional observations into structured representations for efficient learning and inference, while enabling explicit constraints on geometry, topology, and dynamics that are difficult to impose in raw sensor space. Advances in representation learning, discrete latent modeling (tokenization), and vision-language pretraining have added new affordances: discrete latents enable tokenized generative modeling and sequence-based reasoning~\cite{oord2017vqvae}, and large-scale vision-language pretraining with language/semantic supervision provides transferable semantic grounding signals that can guide or distill reasoning-related representations~\cite{radford2021clip}. Modern world models for automated driving increasingly exploit these capabilities to move beyond pixel-level or perceptual reconstruction toward decision-relevant rollout interfaces~\cite{wang_drivedreamer_2025, li_think2drive_2024, min_driveworld_2024, yang_worldrft_2025, yang_raw2drive_2025}.

Despite rapid progress, the literature remains fragmented across task-centric and architecture-centric views, such as forecasting versus planning, Diffusion models versus Transformer generators, open-loop prediction versus closed-loop control. This fragmentation obscures shared internal mechanisms that drive success or failure. Many empirical gaps, including long-horizon drift and hallucination, cross-domain generalization failures, as well as mismatches between perceptual metrics and closed-loop safety, can be traced back to how latent representations encode structure, evolve over time, and align with semantics, objectives, and computational constraints. A latent-centric perspective provides a principled basis to compare methods that otherwise appear disjoint, and clarifies why improvements in visual fidelity or open-loop accuracy do not necessarily translate into safer closed-loop behavior.

\begin{figure*}[t]
\centering
\includegraphics[width=1.0\textwidth]{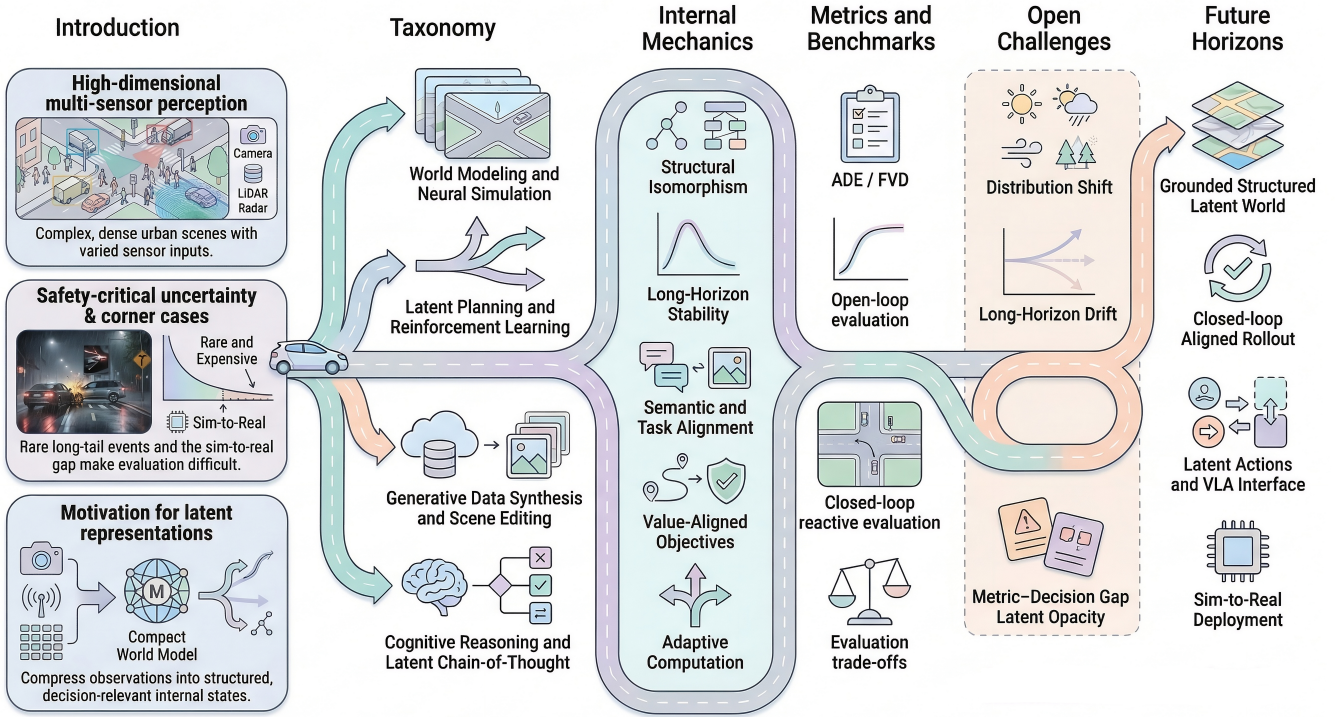}

\caption{\textbf{A visual roadmap visualization of the paper.} }
\label{fig:roadmap}
\end{figure*}

\begin{figure*}[t]
\centering
\includegraphics[width=1.0\textwidth]{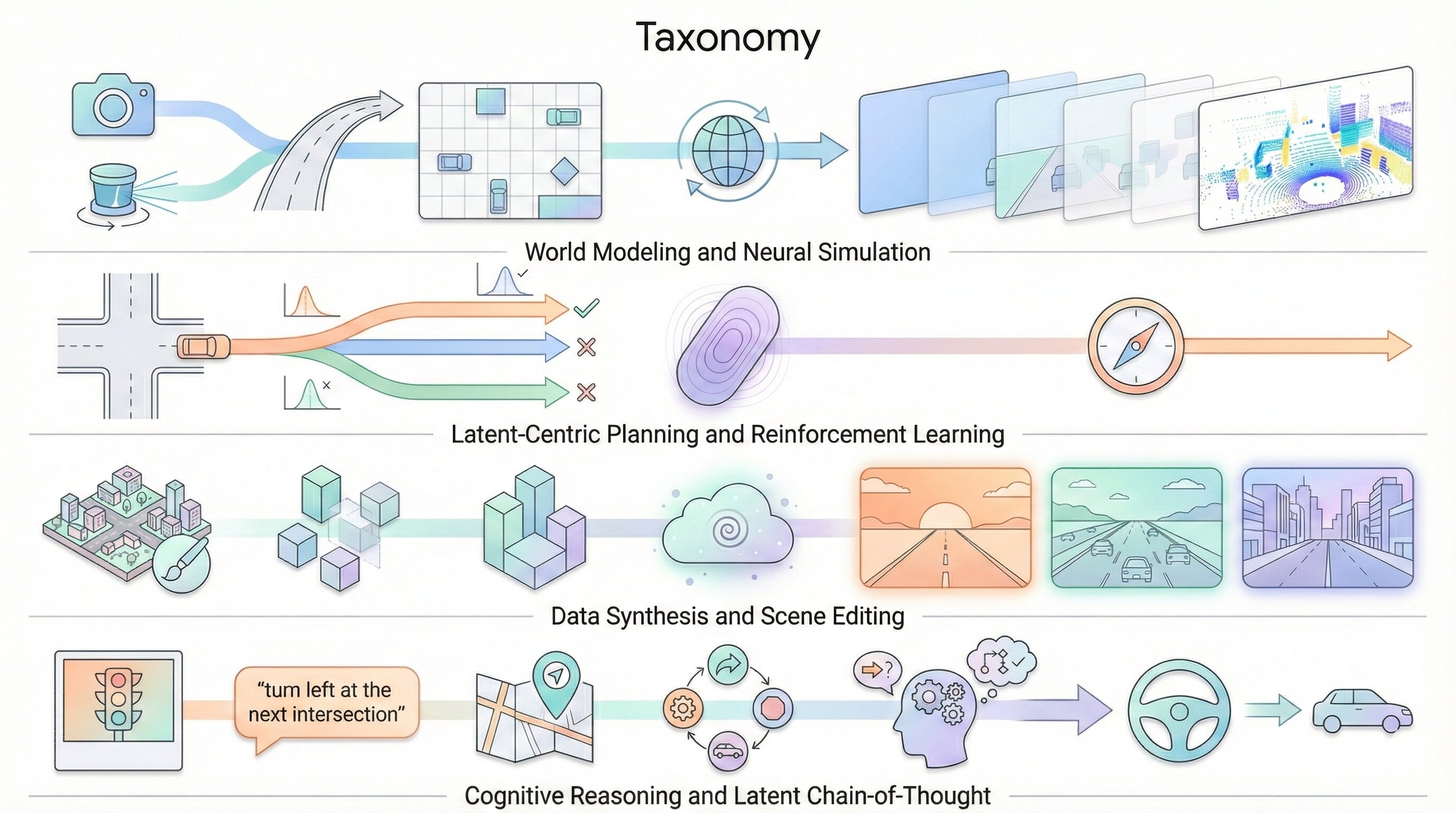}
\caption{\textbf{Taxonomy of world models for automated driving}: A conceptual overview of \textit{Neural Simulation}, \textit{Latent Planning}, \textit{Data Synthesis}, and \textit{Cognitive Reasoning} within a unified latent-centric framework.
}
    \label{fig:taxonomy}
\label{fig:overall_structure}
\end{figure*}

To address these fragmentations, the paper adopts a unifying latent-space perspective and taxonomy that reframes recent progress in world models for automated driving through the lens of representation design and decision alignment. Rather than organizing methods purely by task or architecture, this paper emphasizes the structural and computational properties of latent representations that directly influence closed-loop deployment. Fig.~\ref{fig:roadmap}, which serves as a conceptual “virtual journey” rather than a literal process pipeline, unfolds the paper as a coherent design roadmap: it begins with the practical complexities and challenges in raw sensory perception and learning based approaches (as illustrated in \textit{Section I Introduction}), develops a compact taxonomy of modeling paradigms (\textit{Section II Taxonomy}), and then converges to highlight core internal mechanics that govern rollout behaviour and stability (\textit{Section III Internal Mechanics}). Building on this foundation, the paper examines how evaluation standards differ between open-loop prediction and closed-loop interaction (\textit{Section IV Metrics \& Benchmark}), and navigates through unresolved open challenges (\textit{Section V Open Challenges}), and finally outlines future directions centered on structured latent representations and unified evaluation principles (\textit{Section VI Future Horizons}).

In short, the main contributions of this paper are fivefold: (1) it introduces a unifying taxonomy that characterizes the targets, forms, and structural priors used by modern world models for automated driving and situates them relative to downstream decision interfaces; (2) it synthesizes five cross-cutting internal mechanics (i.e., structural isomorphism and geometric priors, long-horizon temporal dynamics, semantic and reasoning alignment, value-aligned objectives and post-training, and adaptive computation for deliberative rollouts) and analyzes how these mechanics affect closed-loop robustness, generalization, and deployability; (3) it proposes a compact evaluation prescription, comprising a closed-loop metric suite and a resource-aware deliberation cost, designed to reduce the gap between open-loop perceptual scores and closed-loop safety-critical outcomes; (4) it derives prescriptive design recommendations and a prioritized research agenda for progressing from visually plausible rollouts to decision-relevant, verifiable models; and (5) it summarizes and assembles representative benchmarks and exemplar methods to facilitate reproducible follow-up studies. Collectively, these contributions shift the paper from cataloguing to a prescriptive design and evaluation agenda aimed at accelerating the translation of latent world models into deployable automated driving systems.

\section{Taxonomy of World Models for Automated Driving}
\sloppy{}
Fig.~\ref{fig:taxonomy} presents a conceptual overview of the proposed taxonomy, showing how distinct modeling paradigms connect through a shared latent-space perspective. Complementing this overview, Table~\ref{tab:comprehensive_comparison} presents a compact, attribute-level comparison of representative methods and studies, organized by latent representation form, modeling objective, and integration with downstream decision tasks. Together, the figure and table instantiate a unified view that links neural simulation, latent-centric planning, generative data synthesis and scene editing, as well as cognitive reasoning and latent chain-of-thought mechanisms within the world modeling frameworks. Rather than treating these paradigms as isolated directions, the taxonomy highlights shared structural choices and trade-offs, such as representation manifold (continuous, discrete, or hybrid), imposed geometric or topological priors, temporal factorization strategies, and value-aligned objectives, that critically influence rollout fidelity, reasoning capacity, and suitability for closed-loop control. The remainder of this section elaborates on these dimensions in detail, clarifying how different paradigms instantiate and balance these underlying design principles.

\subsection{Spatiotemporal World Modeling and Neural Simulation}
This category encompasses methods that construct neural simulators to approximate the physical world's evolution, focusing on generating high-fidelity, spatiotemporally consistent future observations, ranging from 2-dimensional (2D) video sequences to 3-dimensional (3D) occupancy flows, to serve as the foundation for downstream perception and prediction tasks. Here, \emph{neural simulation} refers to action-conditioned rollouts in a learned latent dynamics model for forecasting and decision making, rather than photorealistic rendering.

To resolve geometric discrepancies between modalities, \textbf{BEVWorld}~\cite{zhang_bevworld_2025}
projects heterogeneous inputs (e.g., from LiDAR and camera) into a unified BEV latent space, enabling joint self-supervised reconstruction via ray-casting. Taking multimodal unification further, \textbf{OmniGen}~\cite{tang_omnigen_2025} addresses the misalignment in sensor synthesis by leveraging a shared BEV space to jointly decode LiDAR and camera data via volume rendering, ensuring strict geometric consistency between synthesized point clouds and multi-view images. To enhance generalization, \textbf{DriveX}~\cite{shi_drivex_2025} introduces a decoupled modeling strategy that separates world representation learning from future state decoding, utilizing dynamic-aware ray sampling to capture holistic scene evolution.

Maintaining consistency in long-horizon predictions remains a primary challenge due to autoregressive error accumulation. \textbf{Epona}~\cite{zhang_epona_2025} addresses this via spatiotemporal factorization, decoupling temporal dynamics from fine-grained spatial generation. Addressing the discrete-continuous debate, \textbf{Orbis}~\cite{mousakhan_orbis_2025} demonstrates that continuous autoregressive flow matching~\cite{lipman2022flow} outperforms discrete tokenization in long-term stability without requiring depth or map supervision. Furthermore, \textbf{LongDWM}~\cite{wang_longdwm_2025} employs cross-granularity distillation, using fine-grained flows to supervise coarse-grained latent dynamics for infinite video generation.

For consistent surround-view synthesis, \textbf{UniMLVG}~\cite{chen_unimlvg_2025} integrates explicit viewpoint modeling into diffusion Transformers, while \textbf{CoGen}~\cite{ji_cogen_2025} utilizes generated 3D geometric conditions to guide the video generation process. Extending beyond video, \textbf{DriveWorld}~\cite{min_driveworld_2024} and \textbf{Drive-OccWorld}~\cite{yang_driving_2025} leverage decoupled memory banks (separating static and dynamic features) to predict 4-dimensional (4D) occupancy flows, effectively bridging generative simulation with downstream perception and planning tasks.

\begin{table*}
\centering
\label{tab:comprehensive_comparison}

% 2. 将 Caption 放在表格主体下方
\caption{\textbf{Comparison of Representative World Model Methods: paradigms and attributes}}
\label{tab:comprehensive_comparison}

\resizebox{\textwidth}{!}{%
\begin{tabular}{l|l|c|l|l}
\toprule
\textbf{Method Name Abbreviation} & \textbf{Latent Structure} & \textbf{Latent Dynamics Model} & \textbf{Key Mechanism} & \textbf{Input Source $\to$ Output} \\
\midrule
\multicolumn{5}{c}{\textbf{Part 1: Neural Simulation \& World Modeling}} \\
\midrule
\textbf{BEVWorld}~\cite{zhang_bevworld_2025} & Multimodal BEV Grid & Transformer & Multimodal Tokenization via Ray-casting Rendering and Diffusion & Camera, LiDAR $\to$ Point Cloud, Video \\
\textbf{CoGen}~\cite{ji_cogen_2025} & 2D + 3D Conditions & Diffusion & 3D Consistent Video Generation via Adaptive Geometric Conditioning & Camera $\to$ Video \\
\textbf{Drive-OccWorld}~\cite{yang_driving_2025} & 3D Occupancy & Transformer & 4D Occupancy Forecasting with Semantic and Motion Normalization & Camera $\to$ Occupancy, Flow \\
\textbf{DriveDreamer}~\cite{wang_drivedreamer_2025} & 2D Latent & Diffusion & Two-Stage Training for Structured Traffic Constraints Comprehension & Camera, 3D Box $\to$ Video \\
\textbf{DriveScape}~\cite{wu_drivescape_nodate} & 2D + 3D Road Prior & Transformer & Bi-Directional Modulated Feature Fusion for Multi-View Synthesis & Camera $\to$ Video \\
\textbf{DriveWorld}~\cite{min_driveworld_2024} & Dual Memory & MSSM & Spatiotemporal Memory State-Space Modeling with Dynamic-Static Separation & Camera $\to$ 4D Features \\
\textbf{DriveX}~\cite{shi_drivex_2025} & Decoupled & Transformer & Omni Scene Modeling with Dynamic-Aware Ray Sampling & Camera, LiDAR $\to$ 3D Representation \\
\textbf{Epona}~\cite{zhang_epona_2025} & Factorized & Diffusion & Decoupled Spatiotemporal Factorization in Autoregressive Latent Space & Camera $\to$ Long Video \\
\textbf{GAIA-2}~\cite{russell_gaia-2_2025} & 2D Latent & Diffusion & Controllable Scene Synthesis via Multi-Agent Latent Embeddings & Camera, Text $\to$ Video \\
\textbf{GLAD}~\cite{xie_glad_2025} & 2D Latent & Diffusion & Streaming Scene Generation via Latent Variable Propagation & Camera $\to$ Streaming Video \\
\textbf{LongDWM}~\cite{wang_longdwm_2025} & Hierarchical & Transformer & Cross-Granularity Motion Distillation for Infinite Video Generation & Camera $\to$ Infinite Video \\
\textbf{MagicDrive-V2}~\cite{gao_magicdrive-v2_2025} & 2D Latent & Diffusion & Spatiotemporal Conditional Encoding for Adaptive Geometry Control & Camera, Text $\to$ Video \\
\textbf{Orbis}~\cite{mousakhan_orbis_2025} & 1D Vector & Flow Matching & Continuous Autoregressive Flow Matching for Long-Horizon Stability & Camera $\to$ Video \\
\textbf{UniMLVG}~\cite{chen_unimlvg_2025} & 2D Latent & Diffusion & Explicit Viewpoint Modeling for Motion Transition Consistency & Camera $\to$ Multi-View Video \\
\midrule
\multicolumn{5}{c}{\textbf{Part 2: Latent-Centric Planning \& Reinforcement Learning}} \\
\midrule
\textbf{AD-R1}~\cite{yan2025adr1closedloopreinforcementlearning} & Latent State & RL & Closed-loop reinforcement learning for end-to-end driving policy optimization & Camera $\to$ Driving Action \\
\textbf{BRYANT}~\cite{wang_autonomous_2025} & Frequency Domain & Transformer & Brain-Inspired Causality-Aware Contrastive Learning with CfC Network & Camera $\to$ Driving Action \\
\textbf{DriveLaW}~\cite{xia_drivelawunifying_2025} & Video Latent & Diffusion & Unified Video Generation and Planning via Direct Latent Injection & Camera $\to$ Trajectory, Video \\
\textbf{EGADS}~\cite{tang_efficient_2026} & 1D Vector & Flow + VAE & Variational Inference with Normalizing Flows for Sample Efficiency & Camera $\to$ Driving Action \\
\textbf{GenAD}~\cite{zheng_genad_2025} & Instance Tokens & Transformer & Map-Aware Instance-Centric Scene Tokenization and Prior Modeling & Camera $\to$ Planning Trajectory \\
\textbf{Interp-E2E}~\cite{chen_interpretable_2022} & 1D + Mask & RL & Sequential Latent Explanation via Semantic Bird-Eye Masking & Camera $\to$ Driving Action \\
\textbf{LatentDriver}~\cite{xiao_learning_2025} & 1D Vector & Transformer & Probabilistic Mixture Distribution for Uncertainty-Aware Decision Making & Camera $\to$ Planning Trajectory \\
\textbf{LAW}~\cite{li2024enhancing} & 2D Latent & Transformer & Self-Supervised Future Scene Feature Prediction for End-to-End Planning & Camera $\to$ Planning Trajectory \\
\textbf{MomAD}~\cite{song_dont_nodate} & 1D Vector & Transformer & Momentum-Aware Planning with Topological Trajectory Matching & Camera $\to$ Planning Trajectory \\
\textbf{Raw2Drive}~\cite{yang_raw2drive_2025} & 1D Vector & Transformer & Guidance Mechanism for Aligned Privileged-to-Raw Model Distillation & Camera $\to$ Driving Action \\
\textbf{SEM2}~\cite{gao_enhance_2024} & Masked Latent & Transformer & Semantic Masking and Recurrent Filtering for Robustness & Camera $\to$ Driving Action \\
\textbf{Think2Drive}~\cite{li_think2drive_2024} & 1D Vector & RSSM & Neural World Model Simulator for Efficient Reinforcement Learning & Camera $\to$ Steering Action \\
\textbf{World4Drive}~\cite{zheng_world4drive_2025} & 2D Feature & Transformer & Intention-Aware Alignment via Vision Foundation Model Priors & Camera $\to$ Planning Trajectory \\
\textbf{WorldRFT}~\cite{yang_worldrft_2025} & 1D Vector & Transformer & Reinforcement Fine-Tuning (RFT) via Group Relative Policy Optimization & Camera $\to$ Trajectory \\
\midrule
\multicolumn{5}{c}{\textbf{Part 3: Generative Data Synthesis \& Scene Editing}} \\
\midrule
\textbf{LiDAR-EDIT}~\cite{ho_lidar-edit_2025} & BEV Tokens & MaskGIT & Spherical Voxelization for Projective Geometry Consistency in Layout Editing & Point Cloud $\to$ Edited Point Cloud \\
\textbf{OmniGen}~\cite{tang_omnigen_2025} & Shared BEV & DiT + ControlNet & Unified Multimodal (LiDAR+Cam) Generation via Volume Rendering & Camera, LiDAR $\to$ Sensor Data \\
\textbf{Safety-Critical}~\cite{peng_safety-critical_2025} & Graph Latent & Diffusion & Adversarial Guidance Sampling for Physically Plausible Traffic Simulation & Trajectory $\to$ Safety-Critical Trajectory \\
\textbf{SubjectDrive}~\cite{huang_subjectdrive_2025} & 2D Latent & Diffusion & Subject Control Mechanism for Scaling Generative Data Diversity & Camera $\to$ Synthetic Image \\
\textbf{SynDiff-AD}~\cite{goel_workshop_nodate} & 2D Latent & Diffusion & Subgroup-Targeted Synthetic Data Augmentation via Semantic Prompts & Text $\to$ Synthetic Image \\
\midrule
\multicolumn{5}{c}{\textbf{Part 4: Cognitive Reasoning and Latent Chain-of-Thought}} \\
\midrule
\textbf{Alpamayo-R1}~\cite{nvidia_alpamayo-r1_2025} & 1D Vector & Transformer & Chain of Causation Reasoning with Reinforcement Learning Feedback & Camera $\to$ Planning Trajectory \\
\textbf{ColaVLA}~\cite{peng_colavla_2025} & Meta-Action & Transformer & Hierarchical Planning via Cognitive Meta-Action Embeddings & Camera, Text $\to$ Trajectory \\
\textbf{CRiTIC}~\cite{ahmadi_curb_2025} & Graph Latent & Transformer & Causal Discovery with Attention Gating for Robust Prediction & Camera $\to$ Agent Trajectory \\
\textbf{Exp-Latent}~\cite{bairouk_exploring_2024} & VAE Latent & - & Automatic Latent Perturbation for Probing Interpretability & Camera $\to$ Steering Command \\
\textbf{FutureSight}~\cite{zeng_futuresightdrive_2025} & Video Latent & Diffusion & Spatiotemporal Visual Chain-of-Thought for Future State Prediction & Camera $\to$ Planning Trajectory \\
\textbf{FutureX}~\cite{lin_futurex_2025} & Latent CoT & Transformer & Dynamic "Auto-think Switch" for Adaptive Reasoning vs. Reflex & Camera $\to$ Trajectory \\
\textbf{LCDrive}~\cite{tan_latent_2025} & Latent Actions & Transformer & Latent Chain-of-Thought (CoT) with Action-World Token Interleaving & Camera $\to$ Action Tokens \\
\textbf{MindDrive}~\cite{sun_minddrive_2025} & 1D Vector & VLM & ``What-if" Simulation with VLM-based Multi-Objective Evaluation & Camera $\to$ Trajectory \\
\textbf{OmniDrive-R1}~\cite{zhang_omnidrive-r1_2025} & Visual Tokens & Transformer & Reinforcement-Driven Visual Grounding (Clip-GRPO) for CoT & Camera, Text $\to$ Action \\
\textbf{VA-VAE}~\cite{yao_reconstruction_nodate} & Visual Token & - & Vision Foundation Model Alignment for Taming Optimization Dilemmas & Camera $\to$ Image Reconstruction \\
\textbf{VERDI}~\cite{feng_verdi_2025} & 1D Vector & - & Vision-Language Model Reasoning Distillation into Modular AD Stacks & Camera $\to$ Action, Reasoning Text \\
\textbf{Vis-Ling}~\cite{shenkut_visual-linguistic_2025}& 1D Vector & Transformer & Multimodal Reasoning for Visual-Linguistic Semantic Feature Fine-tuning & Camera $\to$ Pedestrian Trajectory \\
\textbf{VLA-R}~\cite{noauthor_vla-r_nodate} & Embedding Space & Retrieval & Open-World Vision-Language Action Retrieval Paradigm & Camera, Text $\to$ Driving Action \\
\bottomrule
\end{tabular}%
}
\vspace{5pt} % 增加一点垂直间距，让表格和标题不那么挤

\end{table*}

\subsection{Latent-Centric Planning and Reinforcement Learning}
Moving beyond pixel-level processing, latent-centric planning approaches leverage the compressed and structured latent spaces of world models to facilitate efficient trajectory planning and policy learning. By exploiting the ``dreaming" capability of these models, they simulate future outcomes to optimize decision-making strategies without the computational overhead of high-dimensional sensory data.

Traditional end-to-end automated driving approaches often rely on cascaded pipelines following the sequence of ``perception $\to$ prediction $\to$ planning''. These systems struggle to explicitly capture the interactive future evolution between traffic participants and lack structured prior modeling of trajectory distributions. To bridge this gap, \textbf{GenAD}~\cite{zheng_genad_2025} reframes automated driving as a problem of ``future generation", and unifies the modeling of both agent and ego-vehicle motion evolution within a shared latent space, generating multimodal future trajectory distributions via latent sampling. This formulation enables simultaneous multi-vehicle motion prediction and ego-vehicle planning within a single, cohesive framework. Taking unification a step further, \textbf{DriveLaW}~\cite{xia_drivelawunifying_2025} dissolves the boundary between video generation and motion planning. Unlike cascaded pipelines, it directly injects the latent representation from a high-fidelity video generator into the planner, ensuring that the planned trajectory is intrinsically consistent with the forecasted future dynamics.

While effective, autoregressive world models often grapple with representing multi-modal decision-making and suffer from ``self-delusion" during closed-loop planning. Addressing these issues, \textbf{LatentDriver}~\cite{xiao_learning_2025} models the transition to the next state and the ego-vehicle action as a mixture distribution, from which deterministic control is derived. Furthermore, it mitigates self-delusion by feeding sampled actions from intermediate layers back into the world model, ensuring more robust long-horizon predictions.

A distinct advantage of latent world models is their role as neural simulators for reinforcement learning (RL), which significantly accelerates training efficiency and reduces sample complexity. \textbf{Think2Drive}~\cite{li_think2drive_2024} acts as a neural simulator that learns environmental transfers in a low-dimensional latent space. This architecture enables the planner to conduct extensive parallel ``dream" rollouts, thereby achieving expert-level driving strategies with high training efficiency. Similarly, \textbf{EGADS}~\cite{tang_efficient_2026} leverages RL agents to compress historical information into the latent space using variational inference coupled with normalizing flows. This probabilistic approach facilitates the effective extraction of driving-relevant historical features while minimizing sample complexity.

However, applying model-based RL (MBRL) directly to raw sensor data often faces challenges such as unstable training, distribution drift during long-term rollouts, and the lack of explicit state information. \textbf{Raw2Drive}~\cite{yang_raw2drive_2025} addresses these by proposing a dual-stream MBRL framework. It first trains a world model and planner using privileged information to establish stable priors. Subsequently, it employs alignment mechanisms to synchronize the raw-sensor branch (which is trained solely on multi-view images) with the privileged branch. This paradigm ensures that the policy learned from raw inputs inherits the robustness of privileged knowledge, achieving superior generalization in closed-loop end-to-end configurations. To further align latent representations with planning objectives, \textbf{WorldRFT}~\cite{yang_worldrft_2025} introduces ``Reinforcement Fine-Tuning" (RFT). By utilizing Group Relative Policy Optimization (GRPO)~\cite{shao_deepseekmath_2024} with collision-aware rewards, it refines the world model's features to be more planning-oriented, significantly reducing collision rates compared to purely supervised learning.

\subsection{Generative Data Synthesis and Scene Editing}

To mitigate the long-tail distribution challenge in automated driving, studies in this category utilize generative models to synthesize rare safety-critical scenarios or edit existing sensor data, thereby enriching training datasets with diverse, controllable, and physically plausible environments.

Real-world automated driving datasets frequently exhibit severe class imbalance, predominantly featuring common conditions such as ``sunny daytime" while underrepresenting challenging scenarios like ``rainy nights". To rectify this performance-degrading bias, \textbf{SynDiff-AD}~\cite{goel_workshop_nodate} proposes a synthetic data augmentation pipeline leveraging the Latent Diffusion Model \cite{rombach2022high}. By employing ControlNet~\cite{zhang2023addingconditionalcontroltexttoimage} to maintain semantic consistency with original annotation structures, it introduces a ``subgroup-targeted" prompting scheme designed to generate photorealistic images with specific semantic densities and environmental conditions. This strategy yields inherently labelled synthetic data that significantly enhances the robustness of segmentation and end-to-end driving models under long-tail conditions.

While the efficacy of automated driving systems relies on vast amounts of annotated data, the cost of real-world collection remains prohibitive. Furthermore, simply expanding the volume of generated data does not guarantee downstream performance gains; the critical determinant is controllable diversity. \textbf{SubjectDrive}~\cite{huang_subjectdrive_2025} addresses this imperative by introducing a subject control mechanism that enables generative models to incorporate diverse external data sources. This facilitates precise control over subject and content dimensions, enabling the continuous production of diverse, useful, and freely labeled synthetic training data. The study systematically validates that scaling generative data diversity is paramount for enhancing downstream perception models.

Synthesizing 3D sensor data presents additional distinct challenges: generating LiDAR point clouds from scratch often incurs substantial domain gaps, whereas novel view synthesis is limited in producing counterfactual layouts, such as relocating vehicles or altering object configurations, thereby hindering their utility for causal analysis. \textbf{LiDAR-EDIT}~\cite{ho_lidar-edit_2025} circumvents these limitations by first compressing point clouds into discrete BEV latent tokens via VQ-VAE~\cite{oord2018neuraldiscreterepresentationlearning}. Operating within this structured latent space, it employs a MaskGIT-style~\cite{chang2022maskgitmaskedgenerativeimage} iterative prediction mechanism to reconstruct background regions occluded by manipulated objects. This design enables controllable manipulation of object layouts while maintaining geometric coherence and background consistency, thereby supporting realistic and meaningful scene editing.

High-risk interaction data remains equally scarce yet crucial for robust evaluation. \textbf{Safety-Critical}~\cite{peng_safety-critical_2025} targets this deficit by introducing guided diffusion~\cite{dhariwal2021diffusionmodelsbeatgans} within the latent trajectory space. By imposing differentiable safety-critical constraints and adversarial objectives during the sampling phase, the model synthesizes multi-vehicle interaction evolutions that are both directional and physically plausible. This provides a robust solution for generating high-risk, long-tail traffic scenarios while maintaining behavioural fidelity for closed-loop simulation and evaluation.

\subsection{Cognitive Reasoning and Latent Chain-of-Thought}

This emerging paradigm integrates the semantic reasoning capabilities of large Vision-Language Models (VLMs) into the driving stack. However, the field is rapidly shifting from passive VLM-based explanations to active Chain-of-Thought (CoT) reasoning, enabling autonomous agents to transition from intuitive ``System 1" reactors (fast, perception-driven responses) to deliberative "System 2" thinkers (slow, inference-based planning) \cite{kahneman2011thinking} that plan via logical inference. While natural language provides interpretability, it is often computationally inefficient for real-time control. Addressing this bottleneck, \textbf{LCDrive}~\cite{tan_latent_2025} pioneers the concept of ``Latent-CoT", replacing textual reasoning with a sequence of action-aligned latent tokens. By interleaving action-proposal tokens with world-model rollout tokens, it reasons about future outcomes in a compact latent space before committing to a decision. Similarly, \textbf{FutureX}~\cite{lin_futurex_2025} introduces a dynamic ``Auto-think Switch", allowing the agent to autonomously decide whether to enter a computationally intensive ``Thinking Mode" (performing latent CoT rollouts to refine trajectories for complex scenarios) or remain in ``Instant Mode" for routine driving.

To bridge the gap between high-level semantics and low-level control, recent frameworks employ hierarchical abstraction and rigorous grounding mechanisms. \textbf{MindDrive}~\cite{sun_minddrive_2025} establishes a ``what-if" reasoning loop, where a World Action Model generates foresighted candidates that are rigorously critiqued by a VLM-oriented Evaluator across safety and comfort dimensions. \textbf{ColaVLA}~\cite{peng_colavla_2025} further compresses scene understanding into ``meta-action embeddings" (e.g., ``yield", ``bypass") via a Cognitive Latent Reasoner, allowing the subsequent planner to generate causality-consistent trajectories efficiently. Crucially, ensuring that such high-level reasoning remains faithful to sensor inputs requires active alignment. \textbf{OmniDrive-R1}~\cite{zhang_omnidrive-r1_2025} addresses the hallucination problem via an Interleaved multi-modal CoT driven by reinforcement learning. Instead of relying on dense supervision, it utilizes a ``process-based grounding reward" (Clip-GRPO) to enforce real-time consistency between the VLM's textual reasoning and its visual attention focus. This approach complements the ``Chain of Causation" framework in \textbf{Alpamayo-R1}~\cite{nvidia_alpamayo-r1_2025}, which optimizes reasoning traces via RL to ensure that generated plans are logically grounded in physical causal analysis.

For resource-constrained deployment, distilling reasoning capabilities into smaller models remains vital. \textbf{VERDI}~\cite{feng_verdi_2025} embeds VLM reasoning into the latent space of end-to-end models during training, enabling structured commonsense reasoning without inference-time overhead. Diverging from purely textual logic, \textbf{FutureSightDrive}~\cite{zeng_futuresightdrive_2025} proposes a ``Visual Chain-of-Thought", using generated future image frames as intermediate reasoning steps to visualize potential outcomes. Finally, to enhance robustness against irrelevant perturbations, \textbf{CRiTIC}~\cite{ahmadi_curb_2025} incorporates a causal discovery network with attention gating, filtering out non-causal agents that should not influence the ego-vehicle's decision-making.

\section{INTERNAL MECHANICS: STRUCTURE, ALIGNMENT, AND DYNAMICS IN LATENT REPRESENTATIONS}
\sloppy{}

The internal mechanics of latent representations form the backbone of world models for automated driving, governing whether latent rollouts remain physically consistent, temporally stable, and decision-relevant under closed-loop interaction. Rather than treating perception, prediction, planning, and reasoning as separate modules, this section distills \textbf{five cross-cutting mechanisms} that repeatedly emerge across paradigms and largely determine robustness, generalization, and deployability.

\begin{figure*}
    \includegraphics[width=1.0\textwidth]{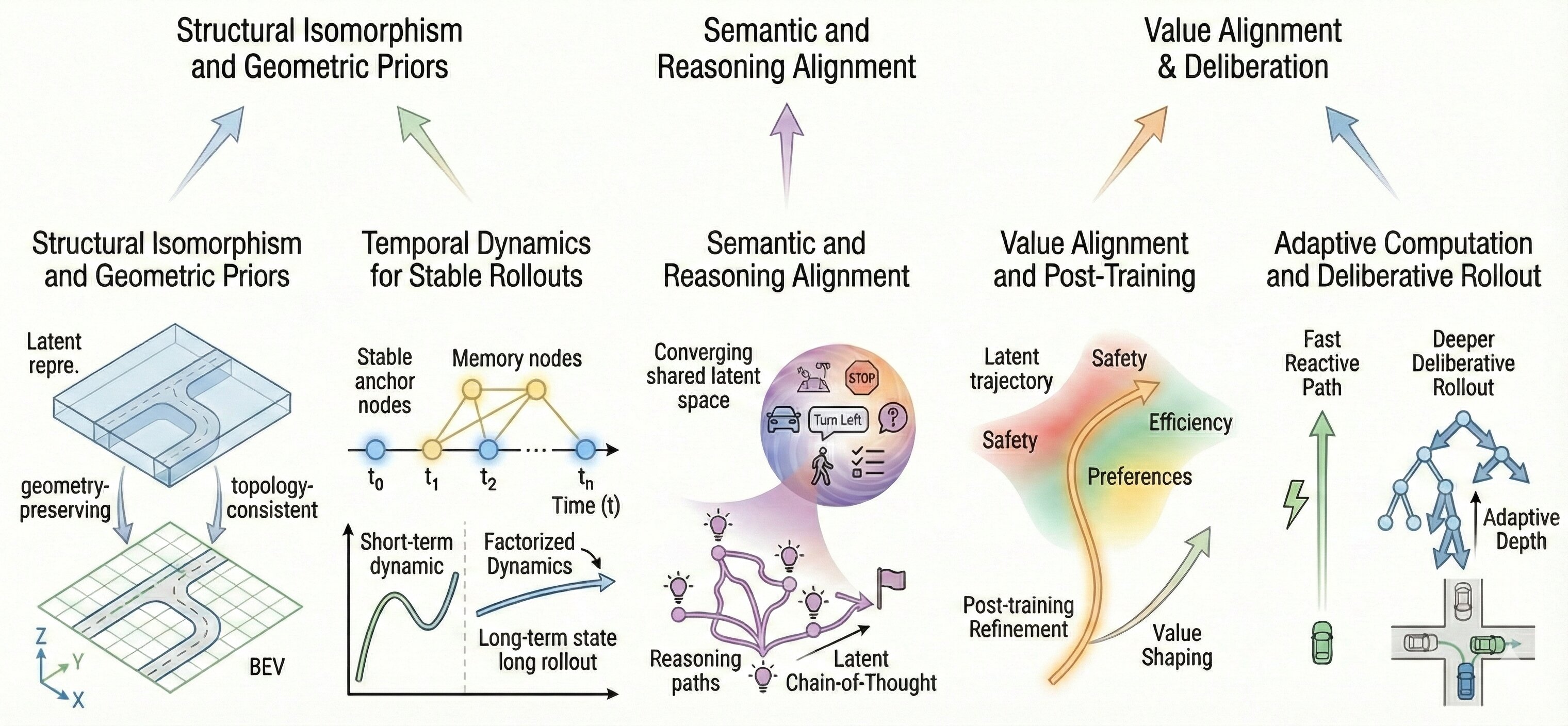}
 \caption{\textbf{Internal mechanics of latent world models for automated driving.}}
    \label{fig:internal_mechanics}
\end{figure*}

As summarized in Fig.~\ref{fig:internal_mechanics}, this section organizes these mechanisms into: \textbf{(A) structural isomorphism and geometric priors} for geometry- and topology-consistent latent structure; \textbf{(B) temporal dynamics and long-horizon stability} to mitigate compounding error during recursive rollout; \textbf{(C) semantic and reasoning alignment} to ground latent variables in transferable abstractions; \textbf{(D) value-aligned objectives and post-training} to couple latent rollouts to safety- and utility-relevant outcomes; and \textbf{(E) adaptive computation and deliberation} to allocate rollout depth under uncertainty and system constraints.

\subsection{Structural Isomorphism and Geometric Priors}
A fundamental paradigm shift in modern world models is the transition from unstructured, low-dimensional latent bottlenecks to spatially isomorphic representations that preserve geometric fidelity. Unlike early approaches that compressed scenes into abstract 1-dimensional (1D) vectors, recent frameworks strictly enforce spatial correspondence within the latent space to retain the physical structure of the driving environment. This structural consistency is primarily pioneered by \textbf{BEVWorld}~\cite{zhang_bevworld_2025}, which organizes the latent space into an explicit Bird’s-Eye-View (BEV) grid rather than a flat vector embedding. This design serves as a strong inductive bias, enabling the decoder to utilize ray-casting rendering for self-supervised learning, thereby ensuring that reconstructed observations maintain rigorous geometric consistency with the physical world. Extending this isomorphism to the volumetric domain, \textbf{Drive-OccWorld}~\cite{yang_driving_2025} incorporates semantic and motion-conditional normalization within a latent occupancy memory to predict the spatiotemporal evolution of occupied voxels via 4D occupancy flows.

Beyond single-modality grids, structural isomorphism has recently emerged as a vital bridge for multi-sensor alignment. \textbf{OmniGen}~\cite{tang_omnigen_2025} demonstrates this by projecting disjoint LiDAR point clouds and multi-view camera features into a unified ``Shared BEV Space''. By enforcing a shared geometric volume, the model ensures that synthesized 3D structures and 2D textures are spatially congruent through a volume rendering decoder, effectively preventing the geometric conflicts inherent in uncoupled generative models. While such grid-based layouts provide dense spatial coverage, researchers have further integrated topological constraints to better model agent-environment interactions. \textbf{GenAD}~\cite{zheng_genad_2025}, for instance, introduces a ``Structural Latent Space'' formulated via map-aware instance tokens. By grounding latent variables to specific map elements and dynamic agents, the framework ensures that sampled trajectories respect the topological structure of the road network, embedding physical admissibility directly into the generative planning process. 

To further constrain the generative solution space and mitigate structural hallucinations, explicit geometric priors are increasingly injected into the latent representation. \textbf{CoGen}~\cite{ji_cogen_2025} demonstrates that high-fidelity 3D geometric conditions are essential for consistent video synthesis, guiding the denoising trajectory beyond simple 2D layout conditions. Adopting a similar philosophy, \textbf{DriveScape}~\cite{wu_drivescape_nodate} fuses 3D road structural priors via Bi-Directional Modulated Transformers to guarantee multi-view consistency. Finally, in the domain of point cloud manipulation, \textbf{LiDAR-EDIT}~\cite{ho_lidar-edit_2025} leverages spherical voxelization as a geometric constraint, enforcing correct projective geometry during the editing of latent tokens to prevent structural distortion in the synthesized output.

\subsection{Temporal Dynamics and Long-Horizon Stability}

Beyond spatial structure, the temporal mechanics of latent representations largely determine whether a world model can support long-horizon imagination without drifting into blur, kinematic inconsistency, or structural hallucinations. A recurring failure mode in autoregressive rollouts is compounding error and temporal collapse, where latent uncertainty and distribution shift accumulate over time and progressively degrade the predicted future. Recent works, therefore, emphasize temporal factorization, memory-structured updates, and training objectives that explicitly narrow the training--inference gap.

Several methods improve long-horizon stability by {structuring} temporal evolution in latent space. \textbf{Epona}~\cite{zhang_epona_2025} advocates decoupled spatiotemporal factorization, separating time-evolving dynamics from fine-grained future generation to reduce autoregressive error accumulation. \textbf{DriveWorld}~\cite{min_driveworld_2024} further formalizes temporal structure via a memory state-space model, distinguishing dynamic entities from static scene content (e.g., dynamic memory banks versus static propagation), thereby imposing different update rules that better reflect physical persistence. \textbf{LongDWM}~\cite{wang_longdwm_2025} addresses the mismatch between short-clip training and long-sequence inference by introducing cross-granularity distillation, using fine-grained motion cues to regularize coarse dynamics and mitigate drift over extended rollouts. From an online generation perspective, \textbf{GLAD}~\cite{xie_glad_2025} adopts a streaming formulation with latent propagation, where the denoised latent from the previous step serves as a prior for the next frame; this design promotes temporal coherence while reducing the need for long-window buffering.

Temporal robustness is also shaped by the \emph{choice of latent manifold} (discrete versus continuous, or hybrid). \textbf{Orbis}~\cite{mousakhan_orbis_2025} provides controlled evidence that discrete token dynamics can be brittle for long-horizon prediction, exhibiting sensitivity to tokenization details, whereas continuous latent dynamics (e.g., flow-matching-based evolution) tend to yield more stable rollouts by avoiding quantization artifacts and enforcing smoother temporal trajectories. Bridging these regimes, \textbf{LCDrive}~\cite{tan_latent_2025} proposes hybrid interleaved dynamics that model world rollout and action proposals within a unified token sequence, leveraging the expressivity of sequence modeling while retaining structured temporal dependencies.

The ability to represent \emph{multi-modality} and avoid overconfident self-consistency further affects long-horizon behavior. \textbf{LatentDriver}~\cite{xiao_learning_2025} models future evolution and ego actions as mixture distributions, enabling diverse hypotheses and reducing self-delusion in long-horizon imagination. Such probabilistic latent dynamics are particularly important when multiple feasible futures exist and deterministic rollouts may collapse to implausible modes.

Mechanisms for intervention and causal analysis help diagnose temporal failures and expose actionable sensitivities. \textbf{Exp-Latent} (short for Exploring Latent Pathways)~\cite{bairouk_exploring_2024} uses automatic latent perturbations (ALP) to probe which latent dimensions influence control outputs, offering a tool to interrogate failure triggers. In parallel, \textbf{BRYANT}~\cite{wang_autonomous_2025} emphasizes frequency-aware temporal disentanglement, separating high-frequency noise-like components from low-frequency trend dynamics to improve robustness and interpretability of temporal evolution. Together, these perspectives suggest that long-horizon capability is not solely a matter of model scale, but of explicitly engineered temporal structure, manifold design, probabilistic dynamics, and diagnostic interventions in the latent space.

\subsection{Semantic and Reasoning Alignment}
To endow the latent space with cognitive reasoning capabilities and physical grounding, modern frameworks increasingly rely on semantic alignment mechanisms that extend beyond simple reconstruction objectives. Instead of learning representations solely from raw sensor data, methods like \textbf{VERDI}~\cite{feng_verdi_2025} introduce a reasoning distillation framework that enforces an alignment during training between intermediate latent features and text embeddings generated by a Vision-Language Model (VLM), effectively internalizing structured reasoning without inference-time overhead.  Building upon this, recent ``System 2" architectures have shifted towards aligning latent dynamics with explicit reasoning traces. \textbf{LCDrive}~\cite{tan_latent_2025} and \textbf{FutureX}~\cite{lin_futurex_2025} advance this paradigm by introducing latent Chain-of-Thought (CoT), where reasoning is represented as a sequence of action-aligned latent tokens that interleave action proposals with world model rollouts to simulate future outcomes. To ensure these internal ``thoughts" remain faithful to the environment, \textbf{OmniDrive-R1}~\cite{zhang_omnidrive-r1_2025} utilizes a reinforcement-driven visual grounding mechanism (Clip-GRPO) that enforces cross-modal consistency between the model's visual attention and its textual reasoning. This is further complemented by \textbf{Alpamayo-R1}~\cite{nvidia_alpamayo-r1_2025}, which aligns latent action predictions with a ``Chain of Causation" to ensure decisions reflect logical causal links rather than mere statistical correlations. 

Beyond linguistic knowledge, the rich visual semantics encapsulated in pre-trained vision foundation models offer powerful inductive biases. \textbf{VA-VAE}~\cite{yao_reconstruction_nodate} aligns the latent space of the visual tokenizer directly with such foundation models to ``tame" the high-dimensional manifold, resolving the optimization dilemma between reconstruction fidelity and generation quality. Similarly, \textbf{World4Drive}~\cite{zheng_world4drive_2025} enriches its representation by injecting spatial-semantic priors for intention-aware planning. This alignment strategy is now expanding towards task-level unification; for instance, \textbf{DriveLaW}~\cite{xia_drivelawunifying_2025} unifies video generation and motion planning by directly injecting latent representations from its generator into a diffusion-based planner. This ensures inherent consistency between high-fidelity future forecasting and reliable trajectory planning, significantly reducing the gap between prediction and control.  Furthermore, \textbf{MindDrive}~\cite{sun_minddrive_2025} and \textbf{ColaVLA}~\cite{peng_colavla_2025} leverage semantic abstractions, such as meta-action embeddings, to align scene understanding with multi-objective decision-making tasks.

In the context of reinforcement learning applied to world models for automated driving, aligning latent representations with ground-truth physics is crucial for bridging the reality gap. \textbf{Raw2Drive}~\cite{yang_raw2drive_2025} addresses this through a privileged alignment strategy, synchronizing a ``raw-sensor branch" with a ``privileged branch" trained on ground-truth actor states.  This concept is extended by \textbf{WorldRFT}~\cite{yang_worldrft_2025}, which introduces Reinforcement Fine-Tuning (RFT) using GRPO to align scene representation learning directly with safety-critical planning objectives. By applying trajectory Gaussianization and collision-aware rewards, WorldRFT refines the latent trajectory space to penalize unsafe behaviors, yielding systematic improvements in zero-shot safety. Finally, semantic alignment facilitates controllable generation. \textbf{SynDiff-AD}~\cite{goel_workshop_nodate} leverages the alignment inherent in Latent Diffusion Models to synthesize specific corner cases based on text descriptions, while \textbf{OmniGen}~\cite{tang_omnigen_2025} utilizes a shared BEV space to align LiDAR and camera generation through volume rendering, ensuring that synthesized multimodal sensor data is geometrically and semantically congruent.

\subsection{Value-Aligned Objectives and Post-Training}
A notable trend in latent world modeling is the transition from reconstruction-centric training to objectives that explicitly reflect {decision utility}. Classical pipelines primarily optimize pixel-level reconstruction or perceptual similarity (e.g., diffusion denoising losses and Fréchet Inception Distance (FID) / Fréchet Video Distance (FVD)-style proxies~\cite{Unterthiner2019FVD}), implicitly assuming that visually faithful rollouts suffice for downstream planning. In safety-critical driving, this assumption is fragile: a rollout can appear sharp yet remain decision-invalid, e.g., by inducing collisions, violating drivable-area constraints, or misrepresenting interactive agent behavior. Recent systems therefore augment or replace purely perceptual objectives with {value-aligned} signals that couple latent rollouts to planning and safety outcomes.

This trend is exemplified by reinforcement-style post-training methods that directly optimize latent representations under task-level rewards. \textbf{WorldRFT}~\cite{yang_worldrft_2025} introduces reinforcement fine-tuning to align scene representation learning with safety-critical planning objectives, using collision-aware rewards to discourage unsafe behaviors in latent trajectory space. In parallel, vision-language-action (VLA)-oriented frameworks increasingly apply post-training to align internal reasoning with reliable action generation: \textbf{OmniDrive-R1}~\cite{zhang_omnidrive-r1_2025} employs reinforcement-driven visual grounding to enforce cross-modal consistency between attention and textual reasoning, while \textbf{Alpamayo-R1}~\cite{nvidia_alpamayo-r1_2025} aligns latent action predictions with a ``Chain of Causation'' to promote logically coherent decision rationales. Value alignment also appears in task-level unification efforts, where the representation learned for forecasting is directly exposed to planning modules; for instance, \textbf{DriveLaW}~\cite{xia_drivelawunifying_2025} injects generator latents into a diffusion-based planner to reduce the prediction--control mismatch.

Value-aligned training introduces new failure modes that are less visible under perceptual metrics. Reward specifications can be sparse or incomplete, creating incentives for reward hacking or over-optimization to proxy notions of safety. Post-training can shift coverage by concentrating probability mass on conservative behaviors, potentially degrading performance in complex interactions. These observations motivate reporting both perceptual quality and decision-centric outcomes, and combining value signals with feasibility constraints and calibrated uncertainty to avoid brittle overconfidence.

\subsection{Adaptive Computation and Deliberation in Latent Rollouts}
Latent world models increasingly function not only as predictive engines but also as computational substrates for planning-time deliberation. Driving requires heterogeneous inference regimes: many situations admit fast reactive control, whereas rare or highly interactive scenarios benefit from deeper lookahead, counterfactual evaluation, and multi-hypothesis reasoning. Latent rollouts enable this spectrum by providing a compact space in which future hypotheses can be generated and scored, while recent VLA/world-model systems suggest that the depth and breadth of such deliberation should be {adaptive} rather than fixed.

Concrete instantiations appear in recent ``reasoning-in-the-loop'' designs that interleave action proposals with world-model imagination. \textbf{LCDrive}~\cite{tan_latent_2025} and \textbf{FutureX}~\cite{lin_futurex_2025} operationalize latent CoT by representing deliberation as sequences of action-aligned latent tokens that alternate between proposing actions and rolling out their consequences, enabling iterative refinement before committing to a control decision. Relatedly, task-level unification approaches such as \textbf{DriveLaW}~\cite{xia_drivelawunifying_2025} support deliberative planning by exposing forecasting latents to the planner, whereas semantic abstraction mechanisms in \textbf{MindDrive}~\cite{sun_minddrive_2025} and \textbf{ColaVLA}~\cite{peng_colavla_2025} provide higher-level latent interfaces that facilitate multi-objective evaluation and candidate selection. Across these designs, adaptive computation is typically realized as variable rollout horizon, branching factor, or refinement iterations, with triggers derived from uncertainty, risk indicators, or disagreement among sampled futures.

Adaptive deliberation raises evaluation and systems questions that remain under-specified in many experimental protocols. Deeper reasoning can improve safety only if it fits within latency, memory, and power budgets; conversely, aggressive early-exit heuristics can fail precisely on rare safety-critical cases. As a result, reporting should couple task metrics with compute budgets (e.g., ms/frame, memory footprint, rollout steps, and branching factor) and characterize how compute allocation affects failure rates. Practical deployment will likely require uncertainty-aware triggers, safety monitors, and graceful fallback behaviors for cases where deliberation is truncated or latent rollouts become unreliable under distribution shift.

\section{Evaluation Standards: Metrics and Benchmarks}
\sloppy{}

\subsection{Open-Loop Fidelity and Closed-Loop Stability}
While high-fidelity sensory reconstruction serves as a foundation for internal simulation, the ultimate utility of a world model is defined by its ability to support safe and robust decision-making. The evaluation of this capability has evolved from static geometric comparisons to dynamic, interaction-aware assessments that examine the model's behavior under uncertainty and error accumulation.

\begin{figure}[t]
    \centering
    \includegraphics[width=\columnwidth]{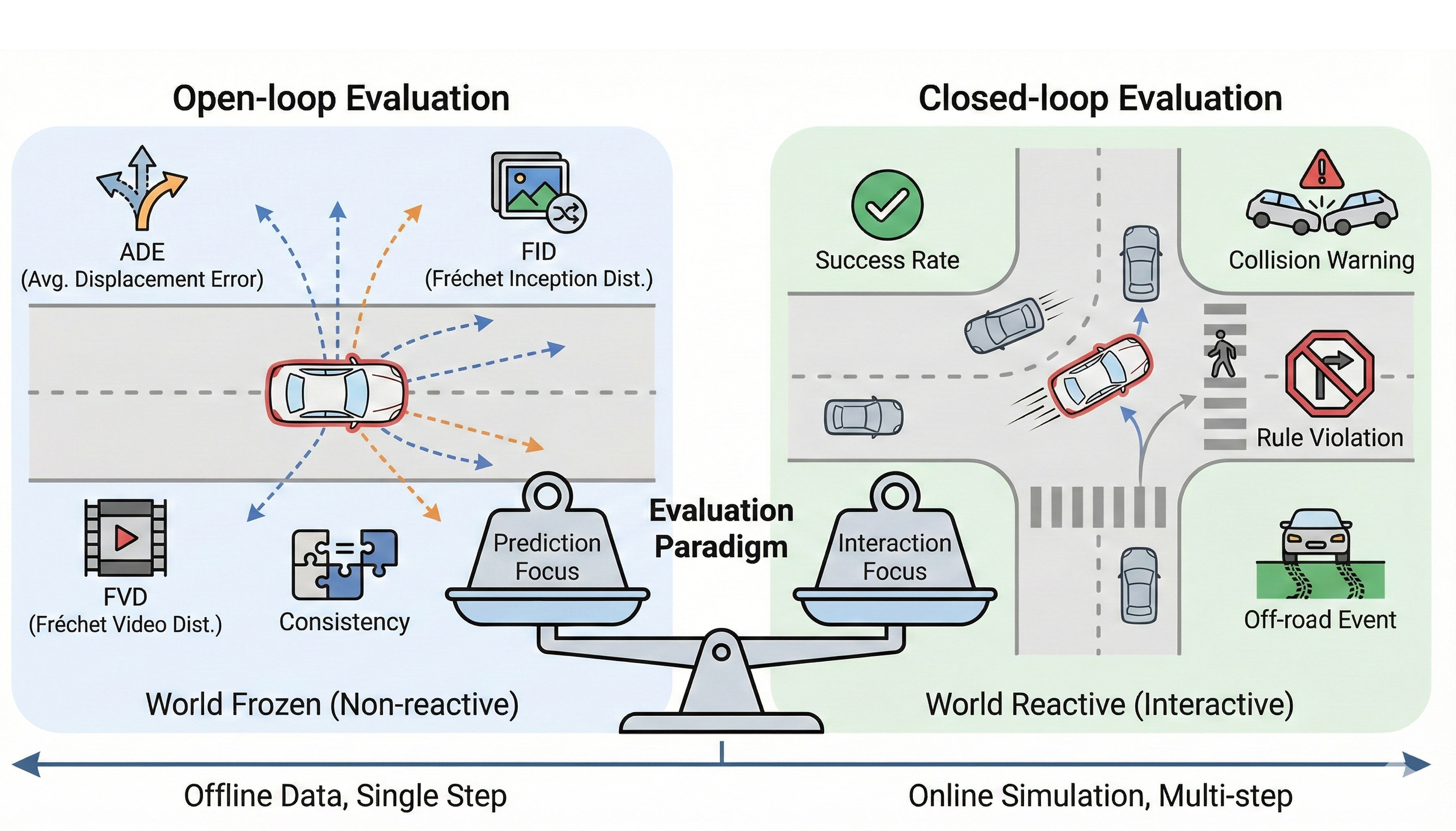}
    \caption{
    \textbf{Evaluation paradigms for latent world models in automated driving}: {open-loop evaluation} (left), and {closed-loop evaluation} (right).
    }
    \label{fig:evaluation_paradigm}
\end{figure}

Fig. \ref{fig:evaluation_paradigm} illustrates the fundamental distinction between open-loop and closed-loop evaluation. Open-loop testing assesses predictive quality in a frozen, non-reactive environment using offline single-step metrics, e.g., Average Displacement Error (ADE),  Fréchet Inception Distance (FID), Fréchet Video Distance (FVD), and consistency. In contrast, closed-loop evaluation measures policy quality in a reactive and interactive world via online multi-step rollouts, focusing on task-level outcomes such as success rate, collision frequency, rule violations, and off-road events. The tension between prediction fidelity and interaction robustness forms the central trade-off in benchmarking latent world models.

The prevailing most common entry point for evaluation remains {open-loop evaluation}, where the model predicts future trajectories based on recorded history. Standard benchmarks on datasets like \textbf{nuScenes} \cite{caesar2020nuscenes} utilize metrics such as ADE and collision rates to quantify the deviation from expert human demonstrations. Baseline methods, including \textbf{GenAD}~\cite{zheng_genad_2025} and \textbf{DriveWorld}~\cite{min_driveworld_2024}, rely on these deterministic measures to validatethe quality of learned representations. Recent advances, such as \textbf{LAW}~\cite{li2024enhancing}, demonstrate that self-supervised latent pre-training significantly reduces these displacement errors compared to planners trained from scratch. Despite its ubiquity, open-loop evaluation introduces a well-known critical ``causal gap''. Since the ego-vehicle's state is reset to the ground truth at every timestamp, the system never experiences the consequences of its own prediction errors or deviations. This masks the ``compounding error'' effects, where minor early drifts push the agent into out-of-distribution states, a failure mode that remains invisible until the system is deployed in an active environment~\cite{li_think2drive_2024}.

To address this limitation, evaluation has shifted toward  {closed-loop protocols} within physics-based interactive simulators like CARLA, where the world model must act as a recursive policy. In this setting, survival metrics such as Success Rate (SR) and Route Completion (RC) become primary indicators. \textbf{Think2Drive}~\cite{li_think2drive_2024} highlights the stark contrast between evaluation paradigms, reporting that models with comparable open-loop errors (e.g., ADEs) can exhibit dramatically different success rates varying from 20\% to 100\% in complex urban scenarios. To capture the nuance of driving quality beyond mere survival, \textbf{Raw2Drive}~\cite{yang_raw2drive_2025} employs a composite Driving Score (DS) that penalizes infractions such as red-light violations and off-road driving, proving that aligning latent states with privileged physical knowledge is essential for adhering to traffic rules, not just avoiding obstacles.

Beyond successful navigation, the temporal stability of the planned trajectory is a critical indicator of latent space robustness. End-to-end models often suffer from ``control jitter'', where plans oscillate wildly between consecutive frames despite a stable environment. Addressing this, \textbf{MomAD}~\cite{song_dont_nodate} proposes the Trajectory Prediction Consistency (TPC) metric to penalize inter-frame inconsistencies, enforcing temporal smoothness in the latent planning process. Complementing this deterministic view, \textbf{LatentDriver}~\cite{xiao_learning_2025} argues for a probabilistic assessment using Negative Log-Likelihood (NLL), ensuring the model captures the multi-modal nature of future possibilities, such as yielding versus bypassing, rather than collapsing into a single, potentially unsafe mean trajectory.

Advanced evaluation protocols now delve deeper into the causal mechanisms driving these decisions, particularly in safety-critical and long-tail scenarios. To distinguish true causal reasoning from spurious correlations, \textbf{BRYANT}~\cite{wang_autonomous_2025} analyzes the latent space in the frequency domain, demonstrating that filtering high-frequency noise reveals the stable causal drivers of the scene. Similarly, \textbf{CRiTIC}~\cite{ahmadi_curb_2025} utilizes causal discovery metrics to verify that the planner's attention aligns with causally relevant agents rather than background distractors. In terms of robustness, \textbf{Alpamayo-R1}~\cite{nvidia_alpamayo-r1_2025} and \textbf{Safety-Critical}~\cite{peng_safety-critical_2025} stress-test models against adversarial perturbations and unstructured terrain, using \textit{Off-Road Rates} to quantify generalization capabilities that standard urban benchmarks fail to capture. Finally, bridging the fidelity-interactivity trade-off, hybrid data-driven simulation environments like \textbf{NAVSIM}~\cite{zhang_epona_2025,zheng_world4drive_2025} now provide large-scale closed-loop-like validation through log-replay rollouts on real-world data, mitigating the simulation-to-reality gap introduced by fully synthetic engines while preserving interactivity.

\subsection{Benchmarks and Simulation Environments}
Reliability in automated driving requires a diverse ecosystem of evaluation platforms, ranging from large-scale static datasets for representation learning to reactive simulators for policy verification. 
Table~\ref{tab:benchmarks_comparison} classifies the prevailing environments (representative datasets and simulation platforms) into three categories based on their fidelity-interactivity trade-off: (1) \textbf{Static Real-World Datasets} for open-loop perception and prediction; (2) \textbf{Interactive Simulation Platforms} for closed-loop policy verification; and (3) \textbf{Data-Driven Neural Simulation Platforms} for hybrid log-simulation.

The \textbf{nuScenes} dataset serves as the standard testbed for open-loop evaluation. Its diverse sensor suite (LiDAR, Camera, Radar) and rich 3D annotations make it the primary choice for baselines like \textbf{GenAD}~\cite{zheng_genad_2025} and \textbf{DriveWorld}~\cite{min_driveworld_2024} to evaluate prediction accuracy (e.g., L2 error) and generation fidelity (e.g., FID). Extensions such as \textbf{nuScenes-Occupancy}~\cite{wang2023openoccupancy} have further enabled voxel-based assessments for methods like \textbf{Drive-OccWorld}~\cite{yang_driving_2025}.  For validating scaling laws, the \textbf{Waymo Open Dataset (WOD)}~\cite{Sun_2020_CVPR}, with its significantly larger volume of driving hours, is utilized by foundation models like \textbf{GAIA-2}~\cite{russell_gaia-2_2025} to ensure robustness across diverse geographies.

To assess closed-loop decision-making, the community relies on \textbf{CARLA}, a high-fidelity open-source simulator based on the Unreal Engine. It serves as the primary arena for MBRL. Notably, recent works like \textbf{Think2Drive}~\cite{li_think2drive_2024} and \textbf{Raw2Drive}~\cite{yang_raw2drive_2025} target the challenging \textbf{CARLA Leaderboard 2.0}~\cite{carlaleaderboard2}, which introduces long-span routes and adversarial scenarios to stress-test the agent's causal reasoning.  While CARLA offers perfect ground-truth physics, it suffers from a visual domain gap compared to real-world sensor streams.

Emerging in 2025, benchmarks like \textbf{NAVSIM}~\cite{dauner2024navsim} bridge the gap between static log-replay and synthetic engines. NAVSIM enables ``Log-Simulation" by allowing the ego-vehicle to deviate from pre-recorded trajectories within real-world logs, synthesizing new views via geometric projection. This paradigm combines the photorealism of real data with the interactivity of closed-loop control. Cutting-edge frameworks, including \textbf{Epona}~\cite{zhang_epona_2025}, \textbf{World4Drive}~\cite{zheng_world4drive_2025}, and \textbf{LAW}~\cite{li2024enhancing}, utilize NAVSIM to validate that world models can learn generalizable policies directly from raw sensor data without the ``gamified" artifacts of traditional simulators.

Beyond CARLA, other interactive simulation platforms such as LGSVL~\cite{rong2020lgsvl}, AirSim~\cite{shah2017airsim}, MetaDrive~\cite{li2022metadrive}, and Waymax~\cite{waymax3666461} can also be adopted to evaluate policy robustness under varying fidelity and scalability constraints, as summarized in Table~\ref{tab:benchmarks_comparison}.

\subsection{Toward Unified Latent-Centric Evaluation Metrics}

Existing benchmarks often decouple open-loop predictive fidelity from closed-loop safety evaluation \cite{bansal2018chauffeurnet, Dosovitskiy2017CARLA,codevilla2018end}. 
While datasets such as nuScenes \cite{caesar2020nuscenes} and WOD \cite{Sun2020Waymo} standardize open-loop metrics, discrepancies between perceptual similarity and interactive robustness remain widely observed. To address this, this paper proposes three complementary unified metrics designed to bridge the gap between latent representation quality and decision-making safety from deployment behavior.

\paragraph{Closed-loop Safety Gap (CSG)}
To quantify the mismatch between visual fidelity and interactive safety, this study defines the Closed-loop Safety Gap:

\begin{equation}
\mathrm{CSG} = F_{OL} - S_{CL},
\end{equation}

\noindent where $F_{OL}\in [0,1] $ denotes a normalized open-loop fidelity score (e.g., inverse ADE or windowed FVD ), and $S_{CL}$ denotes the closed-loop safety score, defined as 

\begin{equation}
S_{CL} = 1 - \frac{\text{Collision Rate}}{\text{Route Complexity}}.
\end{equation}

A large positive CSG indicates that visually plausible predictions fail to translate into safe closed-loop execution, highlighting the long-standing prediction--interaction gap.

\paragraph{Temporal Coherence Score (TCS)}
Stable decision-making further requires smooth inter-frame trajectory evolution. Inspired by prior analyses of control jitter and probabilistic stability \cite{Kendall2017Uncertainty}, this study proposes and defines a normalized Temporal Coherence Score:

\begin{equation}
\mathrm{TCS} = 1 - \frac{\mathrm{Var}(\Delta_t)}{\mathrm{Var}_{\max}},
\end{equation}

\noindent where $\Delta_t$ denotes the frame-to-frame difference of trajectory derivatives.
Higher TCS values indicate reduced temporal oscillation and improved planning stability with smoother and more stable planning dynamics.

\paragraph{Deliberation Cost (DC)}
As world models incorporate multi-step imagination and branching rollouts, evaluation must also consider computational feasibility under real-time constraints \cite{chen2024end}. 
To systematically benchmark the trade-off between cognitive
depth and computational feasibility, this study defines the \emph{Deliberation Cost} ($\mathrm{DC}$) as a normalized efficiency metric that measures the resource footprint required to obtain a marginal safety improvement:

\begin{equation}
\mathrm{DC} = \frac{w_T \,\widehat{T} + w_B \,\widehat{B} + w_E \,\widehat{E} + w_M \,\widehat{M}}{\widehat{\Delta G}}.
\end{equation}

\noindent The numerator summarizes the computational profile of deliberation. Here, $T$ denotes inference latency, measured using a tail statistic (e.g., ms/frame) to reflect worst-case real-time bounds; $B$ is the average branching factor, capturing the breadth of parallel hypotheses or rollouts; $E$ is the energy consumed per decision (J/decision); and $M$ is the working memory footprint (MB). The denominator, $\Delta G$, denotes the expected safety gain attributable to deliberation relative to a purely reactive (zero-deliberation) baseline.

To ensure scale invariance, hatted variables indicate normalization by hardware- and platform-specific reference budgets, e.g., $\widehat{T}=T/T_{\mathrm{ref}}$ and $\widehat{\Delta G}=\Delta G/\Delta G_{\mathrm{ref}}$. Non-negative weights $w_{\{T,B,E,M\}}$ encode deployment priorities; for example, latency may be emphasized in high-speed settings by assigning a larger $w_T$. Lower $\mathrm{DC}$ values indicate more efficient deliberation, encouraging evaluation that jointly accounts for safety benefits and edge-deployable resource constraints.

\begin{table*}[t]
\centering

\caption{\textbf{Evaluation ecosystem for latent world models}: Representative datasets and simulation platforms.}
\label{tab:benchmarks_comparison}

\resizebox{\textwidth}{!}{%
\begin{tabular}{l|l|l|l|l}
\toprule
\textbf{Benchmark Datasets or Platforms} & \textbf{Type} & \textbf{Primary Evaluation Focus} & \textbf{Pros (+) \& Cons (-)} & \textbf{Representative Methods} \\
\midrule
\multicolumn{5}{c}{\textit{Category 1: Static Real-World Datasets (Open-Loop)}} \\
\midrule
\textbf{nuScenes} & Real-world Log & Perception, Generation (FID), Prediction (L2) & 
\textbf{+} Multi-modal standard; Rich annotations \newline 
\textbf{-} Causal gap (No interaction) & 
\textbf{GenAD}~\cite{zheng_genad_2025}, 
\textbf{DriveWorld}~\cite{min_driveworld_2024}, 
\textbf{Drive-OccWorld}~\cite{yang_driving_2025} \\
\midrule
\textbf{Waymo Open Dataset (WOD)} & Real-world Log & Large-scale Scaling, Long-horizon Prediction & 
\textbf{+} High volume for foundation models \newline 
\textbf{-} High computational cost & 
\textbf{GAIA-2}~\cite{russell_gaia-2_2025}, 
\textbf{LatentDriver}~\cite{xiao_learning_2025} \\
\midrule
\multicolumn{5}{c}{\textit{Category 2: Interactive Simulation Platforms (Closed-Loop)}} \\
\midrule
\textbf{CARLA (LB 2.0)} & Physics Engine & Planning Policy, Safety, Rule Compliance & 
\textbf{+} Fully interactive; Reproducible dynamics \newline 
\textbf{-} Visual Sim-to-Real gap & 
\textbf{Think2Drive}~\cite{li_think2drive_2024}, 
\textbf{Raw2Drive}~\cite{yang_raw2drive_2025}, 
\textbf{MomAD}~\cite{song_dont_nodate}, 
\textbf{Alpamayo-R1}~\cite{nvidia_alpamayo-r1_2025} \\
\midrule
\textbf{LGSVL (SVL Simulator)} & Physics Engine & Full AV Stack Integration & 
\textbf{+} High-fidelity sensor models; ROS compatibility \newline 
\textbf{-} Heavy system dependency & -- \\
\midrule
\textbf{AirSim} & Physics Engine & Perception + Control Simulation & 
\textbf{+} Photorealistic rendering; Multi-modal sensors \newline 
\textbf{-} Limited realistic traffic interaction & -- \\
\midrule
\textbf{MetaDrive} & Lightweight Simulator & RL-style Policy Evaluation & 
\textbf{+} Fast and scalable for training \newline 
\textbf{-} Simplified visual realism & -- \\
\midrule
\textbf{Waymax} & Scenario-based Simulation & Log-conditioned Closed-loop Rollout & 
\textbf{+} Large-scale scenario replay \newline 
\textbf{-} Limited physical dynamics modeling & \textbf{LatentDriver}~\cite{xiao_learning_2025},
\textbf{AWMs}~\cite{nachkov2025dream}\\
\midrule
\multicolumn{5}{c}{\textit{Category 3: Data-Driven Neural Simulation Platforms (Hybrid)}} \\
\midrule
\textbf{NAVSIM / OpenDV} & Log-Simulation & Generalizable Policy, End-to-End Planning & 
\textbf{+} Real visuals + Closed-loop freedom \newline 
\textbf{-} Viewpoint limited by sensor bounds & 
\textbf{World4Drive}~\cite{zheng_world4drive_2025}, 
\textbf{Epona}~\cite{zhang_epona_2025}, 
\textbf{LAW}~\cite{li2024enhancing} \\
\bottomrule
\end{tabular}%
}
\vspace{5pt}

\end{table*}

\section{Frontiers and Bottlenecks: Persistent Challenges in Latent World Modeling}
\sloppy{}

Despite rapid progress in latent world models for automated driving, translating strong benchmark results into robust real-world behavior remains challenging. The evaluation standards reviewed in \textit{Section IV} make this gap explicit: open-loop fidelity and short-horizon accuracy often fail to predict closed-loop safety under long-horizon rollouts, domain shift, and interactive feedback. Many failures are therefore not merely engineering artifacts, but stem from how latent representations evolve over time, how computation is allocated during imagination, and how decisions are grounded and verified under uncertainty.

\begin{figure}[t]
    \centering
    \includegraphics[width=\columnwidth]{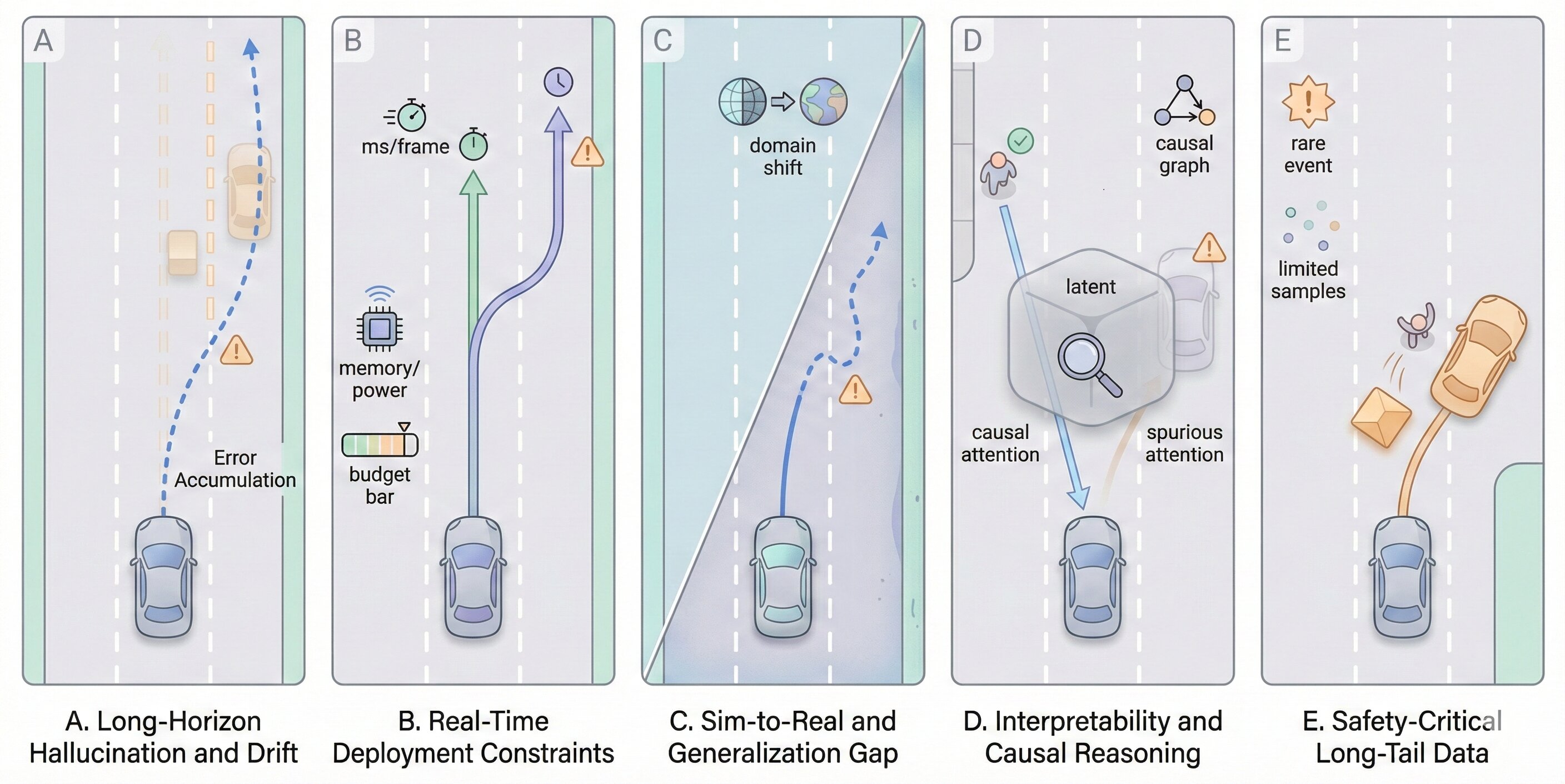}
    \caption{\textbf{Open challenges for latent world models in autonomous driving.}}
    \label{fig:open_challenges}
\end{figure}

Fig.~\ref{fig:open_challenges} summarizes five recurring bottlenecks that organize this section. Panel A highlights long-horizon hallucination and drift from compounding rollout errors; Panel B captures real-time deployment constraints that limit deliberation depth under on-board latency, memory, and power budgets. Panel C emphasizes persistent sim-to-real and cross-domain generalization gaps, while Panel D points to limited interpretability and weak causal reasoning that complicate diagnosis and verification. Panel E further stresses the scarcity of safety-critical long-tail interactions, motivating stress testing and evaluation beyond in-distribution benchmarks. The following subsections discuss each bottleneck in turn and summarize representative mitigation directions.

\subsection{Long-Horizon Consistency and the Hallucination Dilemma}
While recent generative driving world models have made substantial progress in short-horizon visual fidelity, long-horizon rollouts remain fundamentally challenged by stability degradation and hallucination. As the rollout length increases, compounding latent-state errors and distributional shift progressively amplify, resulting in blurred renderings, kinematic drift, and, more critically, structural violations of physical and geometric constraints (e.g., spontaneous object disappearance, multi-view inconsistency, or dynamically implausible ego/agent states). Prior work has begun to clarify both the etiology of these failures and potential mitigation strategies from complementary perspectives. \textbf{Orbis}~\cite{mousakhan_orbis_2025} provides controlled evidence that discrete tokenization can be brittle in long-horizon prediction, exhibiting pronounced sensitivity to tokenizer design, whereas continuous latent dynamics (e.g., flow matching) tend to yield more stable rollouts. \textbf{LongDWM}~\cite{wang_longdwm_2025} attributes long-horizon degradation to the training-inference mismatch between short-clip supervision and long-sequence autoregressive deployment, and alleviates drift via cross-granularity dynamic distillation. \textbf{Epona}~\cite{zhang_epona_2025} addresses autoregressive error accumulation through decoupled spatiotemporal factorization and a chain-of-forward training regime, and further substantiates the decision utility of long-horizon world models on planning benchmarks. Taken together, these results underscore a broader open problem: the field still lacks unified, interpretable, and closed-loop--relevant evaluation protocols for long-horizon stability. A key direction is to establish principled correspondences between long-horizon perceptual metrics (e.g., windowed FVD), geometric consistency, physical feasibility, and closed-loop failure rates (collision/off-road/near-miss), thereby advancing world models from being merely visually plausible to being actionable for decision-making.

\subsection{Real-time Efficiency and Deployment}
Despite rapid advances in diffusion-based generators and large Transformer world models, {real-time} deployment remains a major bottleneck for automated driving: state-of-the-art generative rollouts are often computationally intensive, memory-demanding, and difficult to execute within strict on-board latency and power budgets (e.g., automotive-grade System-on-Chips (SoCs)). Recent works explore complementary directions, yet important limitations persist.

\textbf{GLAD}~\cite{xie_glad_2025} moves diffusion-style driving video generation toward {online} use via frame-by-frame streaming with latent propagation and caching, reducing the need to buffer full sequences and enabling unbounded-length synthesis. Its streaming formulation, however, is still anchored in iterative denoising and may remain prohibitive for hard real-time on-vehicle inference without aggressive acceleration (distillation, few-step sampling, quantization) and careful systems integration; moreover, streaming generation alone does not guarantee closed-loop controllability or safety under interactive feedback.

\textbf{EGADS}~\cite{tang_efficient_2026} addresses efficiency primarily from a {training} perspective, targeting sample complexity and generalization in urban driving by leveraging latent-space belief modeling and demonstration-augmented reinforcement learning. This can reduce data and training cost, yet it does not directly resolve {inference-time} constraints (latency/throughput) for deployment, and its reliance on structured rewards and demonstration signals may limit scalability across domains with different objectives, sensor suites, or driving styles.

\textbf{Think2Drive}~\cite{li_think2drive_2024} improves closed-loop learning efficiency by using a compact latent world model as a neural simulator to perform parallelized rollouts, accelerating policy optimization in complex benchmarks. The approach still incurs the overhead of maintaining a high-capacity world model and a planner, and deployment viability hinges on whether the combined stack can meet real-time constraints and preserve fidelity under distribution shift; simulator-centric training may also leave a residual sim-to-real gap and calls for robust safety monitors for on-road use.

A unifying gap remains in {systems-level} methodologies that jointly optimize (i) rollout/model fidelity, (ii) inference latency and memory footprint, and (iii) closed-loop safety under interactive feedback. Promising directions include low-step generative rollouts (distillation and consistency-type objectives), hardware-aware compression (quantization/pruning), modular scheduling (asynchronous world-model updates with safety-critical fallback), and benchmarks that report both task performance and real-time budgets (ms/frame, memory, power) under realistic automotive constraints.

\subsection{Sim-to-Real Gap and Generalization}
\noindent Generative world models and end-to-end driving policies often exhibit strong in-distribution performance yet degrade sharply under distribution shift, creating a persistent sim-to-real and cross-domain generalization gap. Models trained on a particular log distribution (e.g., a specific city, sensor suite, or weather profile) may fail when deployed in regions with different road topology, traffic norms~\cite{10919561}, sensing characteristics, or long-tail interactions. The difficulty stems from multiple entangled sources of shift, rendering and dynamics mismatches in simulation, sensor noise and calibration differences, behavioral heterogeneity of other agents, and dataset-specific biases, so improvements on a single benchmark rarely translate into robust out-of-domain (OOD) safety.

Recent work explores complementary remedies, each with clear limitations. \textbf{Raw2Drive}~\cite{yang_raw2drive_2025} highlights \emph{privileged alignment} as a way to bridge modality and domain gaps by distilling training-time privileged signals (available in simulation or via richer annotations) into representations learned from raw observations. This strategy can stabilize learning against appearance-level variation, yet its reliance on privileged supervision raises portability concerns: privileged cues may be unavailable, inconsistent, or imperfectly matched to real-world structure, and alignment objectives can be highly sensitive to design choices, reducing reproducibility across domains. \textbf{DriveX}~\cite{shi_drivex_2025} emphasizes {generalizable world knowledge}, arguing for scale and training objectives that encourage transferable structure rather than dataset-specific memorization. Such scaling trends, however, do not by themselves establish that the acquired knowledge is causal or safety-relevant. Generalization claims are often supported by proxy metrics and limited cross-domain protocols, while the resulting models are typically heavier and harder to deploy under real-time constraints. \textbf{World4Drive}~\cite{zheng_world4drive_2025} introduces {foundation-model priors} to improve OOD robustness, leveraging stronger pretrained representations to reduce dependence on narrow training distributions. The benefit is frequently strongest at the perception/representation level, whereas interaction-level decision-making and closed-loop safety can still fail under behavioral and rule shifts. Moreover, incorporating large pretrained priors complicates attribution and verification, making system-level safety arguments more challenging.

Progress in this area calls for evaluation protocols that explicitly stress OOD conditions (new cities, sensor configurations, rare interactions) and report {closed-loop} failure statistics alongside open-loop accuracy. Methodologically, promising directions include disentangling appearance from geometry and dynamics in latent spaces, learning domain-invariant causal factors for interaction modeling, and designing adaptation or calibration mechanisms that are provably safe under shift (e.g., uncertainty-aware fallback policies and safety monitors). Without such protocol and system advances, improvements in generative fidelity or in-distribution planning accuracy are unlikely to translate into reliable real-world deployment.

\subsection{Interpretability and Causal Reasoning}
Learning-based driving policies and world models remain difficult to audit: strong performance can coexist with spurious cues, brittle attention patterns, and post-hoc narratives that are not faithful to the actual decision mechanism. Interpretability methods begin to expose internal sensitivities, yet establishing {causal} and {action-relevant} explanations, with predictive power for failure modes under distribution shift, remains an open challenge.

\textbf{Exp-Latent}~\cite{bairouk_exploring_2024} advances mechanistic interpretability via ALP, probing which latent dimensions drive control outputs and enabling controlled what-if analyses. The resulting explanations, however, are often \emph{model-internal}: latent axes may not align with physically intervenable factors (e.g., agent positions or road geometry), can be unstable across training runs, and typically provide local sensitivity rather than closed-loop guarantees. \textbf{CRiTIC}~\cite{ahmadi_curb_2025} introduces a causal-discovery-driven attention gating mechanism for multi-agent prediction, encouraging attention to track causally relevant entities and improving robustness to non-causal distractions. Its efficacy depends on strong identifiability assumptions from observational data, remains sensitive to hidden confounders and partial observability, and does not directly translate into causal guarantees for planning and control. \textbf{Alpamayo-R1}~\cite{nvidia_alpamayo-r1_2025} pushes toward decision-facing explanations with a ``Chain of Causation'' that couples natural-language causal reasoning to trajectory outputs, offering a path toward human-auditable driving rationales. The approach still faces faithfulness risks (plausible rationalizations versus true drivers of action), evaluation bias when relying on learned critics, and limited coverage of long-tail causal structures.

Progress requires evaluation protocols that test {explanation faithfulness} (whether explanations predict action changes under interventions), link explanations to closed-loop safety outcomes, and support counterfactual analysis at the level of physically meaningful variables. Methodological directions include disentangled and grounded latent factors, causal representation learning with explicit interventions, and safety monitors that exploit causal structure to anticipate failures rather than merely describe them after the fact.

\subsection{Safety-Critical and Long-tail Data}
Safety-critical events (collisions, near-misses, extreme cut-ins, rare rule violations) are exceedingly sparse in real-world driving logs, yielding severe imbalance for both training and evaluation. As a result, models fitted to nominal driving can fail under rare interactions, and standard benchmarks often under-sample the dominant risk modes. Generative augmentation is a natural response, but it raises hard questions about coverage, realism, and safety relevance.

\textbf{Safety-Critical} ~\cite{peng_safety-critical_2025} steers latent diffusion toward hazardous regimes via guidance. Its key limitation is {validity}: guidance can optimize proxy notions of ``danger'' and produce physically or behaviorally implausible scenes, so gains on synthetic distributions may not reduce real-world risk. Coverage remains uncertain because guidance may collapse onto a narrow subset of corner cases.

\textbf{SynDiff-AD}~\cite{goel_workshop_nodate} synthesizes targeted corner cases to enrich long-tail data. Effectiveness hinges on stringent {quality control}, diversity without infeasibility, correct labels, and consistency with sensor physics and traffic rules. Without filtering and closed-loop checks, synthetic data can introduce artifacts, harm calibration, or bias planners; standardized protocols for measuring (i) long-tail coverage, (ii) realism/causal plausibility, and (iii) closed-loop safety impact are still missing.

Stronger benchmarks should report rare-event risk (collision/off-road/near-miss) and support scenario-level validation (rule compliance, physical feasibility, multi-view consistency). Safety-constrained generation, calibrated uncertainty, and simulator-in-the-loop verification are likely prerequisites for converting synthetic long-tail data into measurable closed-loop safety gains.

\section{Future Horizons}
\sloppy{}

Advancing latent world models toward reliable real-world deployment of automated driving requires coordinated progress across model stability, evaluation alignment, representation structure, and system integration. In particular, stabilizing long-horizon rollouts is essential to mitigate compounding errors and hallucinations under distribution shift. At the same time, reconciling open-loop predictive metrics with closed-loop interactive performance is critical to narrowing the long-standing metric–decision gap. Finally, structuring latent spaces for interpretability and task grounding can enable controllable, semantically meaningful reasoning within planning pipelines. 

\begin{figure*}[t]
    \centering
    \includegraphics[width=0.73\textwidth]{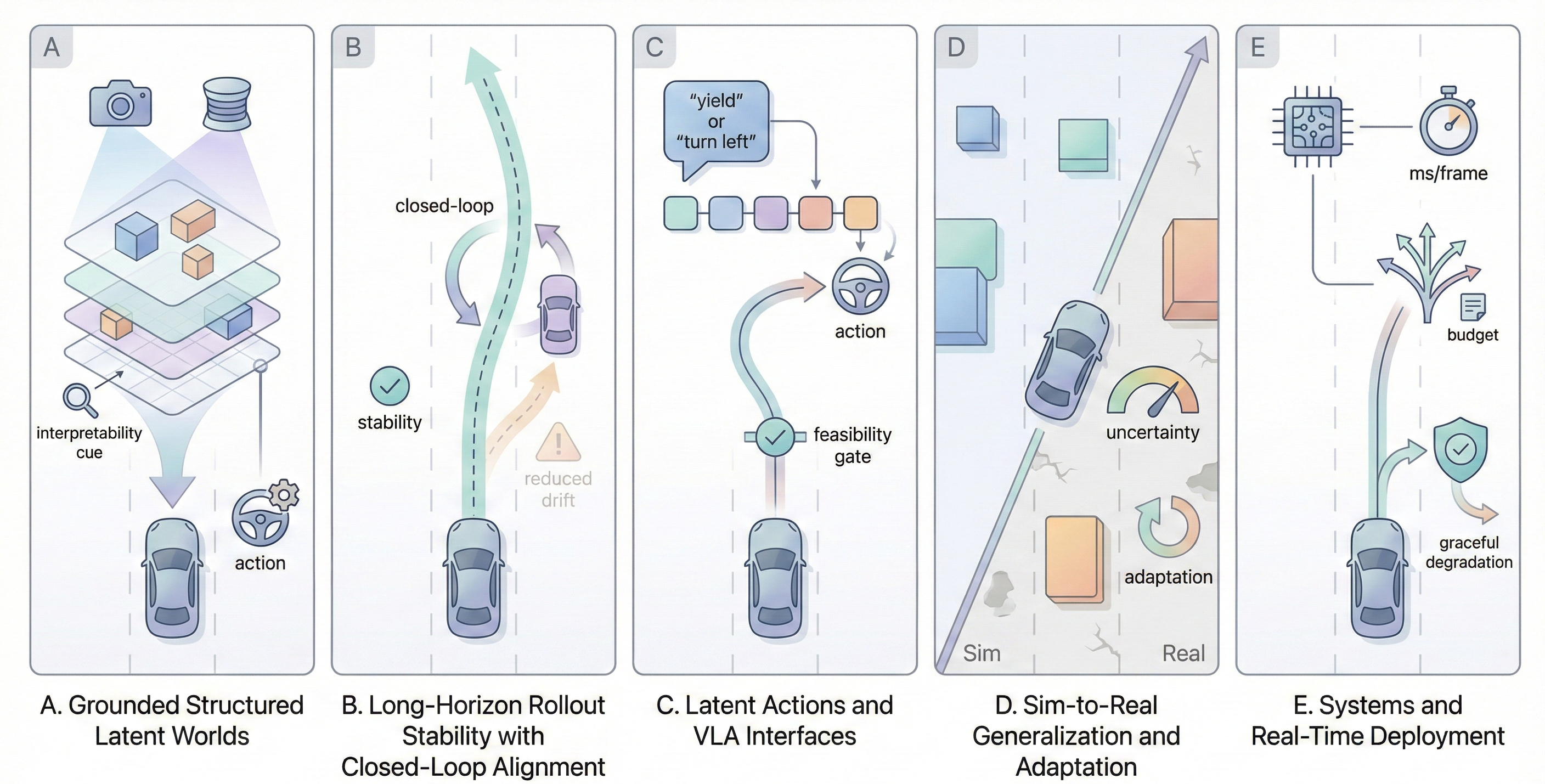}
\caption{\textbf{Illustration of future horizons and research directions for latent world models in automated driving.}
}
    \label{fig:future_horizons}
\end{figure*}

Fig.~\ref{fig:future_horizons} summarizes five complementary research directions toward advancing latent world models for decision-ready, real-world deployable automated driving: 
(A) grounded and structured latent worlds, 
(B) long-horizon rollout stability with closed-loop alignment, 
(C) latent actions and VLA interfaces, 
(D) sim-to-real generalization and adaptation, and 
(E) systems-level optimization for real-time deployment under resource constraints. Collectively, these directions aim to produce world models that are not only perceptually expressive but also causally grounded, computationally efficient, and verifiably safe in interactive driving environments.

\subsection{Grounded and Structured Latent Worlds}
% What
Current latent world models often entangle appearance, geometry, and dynamics, which limits controllability and reliability under distribution shift.
% How
A promising direction is to learn factorized latent states with explicit geometric and kinematic anchors (e.g., BEV/occupancy primitives) while retaining high-fidelity view synthesis. Structured latent representations enable physically meaningful interventions and planning-time constraints. Key challenges persist, which include identifiability of factors, stable cross-view alignment, and evaluation protocols that connect latent structure to closed-loop failure rates.

\subsection{Long-horizon Rollout Stability with Closed-loop Alignment}
Long-horizon rollouts remain susceptible to compounding prediction errors and generative hallucinations; improvements in perceptual metrics do not necessarily imply safer closed-loop behavior. Future research should align training objectives with rollout-time constraints through multi-step supervision, uncertainty-aware rollouts, and simulator-in-the-loop validation that directly penalizes safety-critical failures such as collisions, off-road events, or near-misses. Key open problems include scalable evaluation of rare events, reliable uncertainty calibration, and preventing visually plausible but physically invalid generations.

\subsection{Latent Actions and VLA Interfaces}
Bridging latent world models with VLA-style decision making requires action representations that are both language-groundable and controller-compatible.
Future work can explore hybrid action spaces, combining discrete tokens for semantic grounding with continuous latent variables for control. In such systems, reasoning modules (e.g., large language model (LLM)/VLM-based planners) can generate candidate actions that are validated through feasibility checks in latent world rollouts. Major challenges include standardizing action token representations across platforms and ensuring that language-conditioned policies remain robust under adversarial prompts or OOD contexts.

\subsection{Sim-to-Real Generalization and Adaptation}
Even strong in-distribution models degrade when transferred from simulation to real-world situations, or when operating across different cities, sensor configurations, and traffic norms. Addressing this sim-to-real gap and domain adaptation requires learning domain-invariant causal factors, developing principled privileged-to-raw distillation strategies with explicit validity checks, and introducing test-time adaptation mechanisms gated by calibrated uncertainty and safety monitors. Crucially, such adaptation mechanisms must be verifiably safe and auditable, rather than merely improving average-case benchmark performance.

\subsection{Systems and Real-time Deployment}
Large Diffusion- and Transformer-based world models present significant challenges for automotive latency, memory, and power budgets. Practical deployment therefore requires a mix of model- and systems-level advances: (i) efficient generative rollouts, e.g., few-step/step-reduced diffusion samplers (via distillation or consistency models) and short-horizon/truncated rollout strategies with periodic re-planning; (ii) hardware-aware compression and acceleration, such as quantization, pruning, mixed precision, and kernel-level optimization; and (iii) modular scheduling architectures that permit asynchronous world-model updates while guaranteeing safety-critical fallback behavior. Furthermore, evaluation protocols should report deployment-relevant resource metrics alongside task scores, including latency, memory footprint, energy/power consumption (average and peak), and the hardware platform used. These resource-aware benchmarks let researchers assess not only algorithmic performance but also real-world feasibility.

\section{Conclusion}
\sloppy{}
This paper establishes a unifying latent-space framework for world models in automated driving by synthesizing recent advances through three key perspectives: a latent-centric taxonomy, an analysis of internal model mechanics, and a compact evaluation prescription that includes the proposed CSG/TCS/DC metrics to better align open-loop predictive fidelity with closed-loop safety performance. The analysis reveals a fundamental tension between perceptual realism and closed-loop stability: although current models can generate visually plausible sensor observations, they often fail to guarantee physically consistent and decision-relevant behavior when deployed within interactive control loops.
The five core contributions of this work, namely a taxonomy through a latent-space perspective, a synthesis of internal mechanics, a compact evaluation prescription, prescriptive design recommendations, and a curated set of benchmarks and exemplar methods, collectively provide a structured and deployment-oriented research agenda for translating latent world models into practical automated driving systems.

Looking forward, the convergence of structured geometric priors, language-grounded reasoning, large-scale pretraining with efficient fine-tuning/distillation, resource-adaptive computation, and rigorous closed-loop evaluation will accelerate generalist embodied foundation models for automated driving.
Concretely, these models and methods should unify perception, prediction, and decision-making within a unified and compact latent space, expose verifiable interfaces (calibrated uncertainty and counterfactual queries), as well as support resource-adaptive deliberation with safety-by-design, enabling decision-ready systems that meaningfully narrow the sim-to-real gap. Ultimately, these architectures may form the cognitive backbone for robust, higher-level automated driving, supporting safer and more adaptive autonomy in complex real-world environments.

\bibliographystyle{IEEEtran} 
\bibliography{latent}

\begin{IEEEbiography}[{\includegraphics[width=1in,height=1.25in,clip,keepaspectratio]{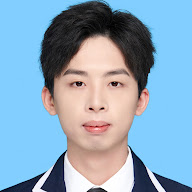}}]{Rongxiang Zeng}
received the B.E. degree from Shenzhen Technology University, Shenzhen, China, in 2025. He is currently pursuing the M.Sc. degree in Robotics Systems Engineering at RWTH Aachen University, Aachen, Germany. His research interests include Robotics, automated driving, world models, and vision-language-action models.
\end{IEEEbiography}

\begin{IEEEbiography} [{\includegraphics[width=1in,height=1.25in,clip,keepaspectratio]{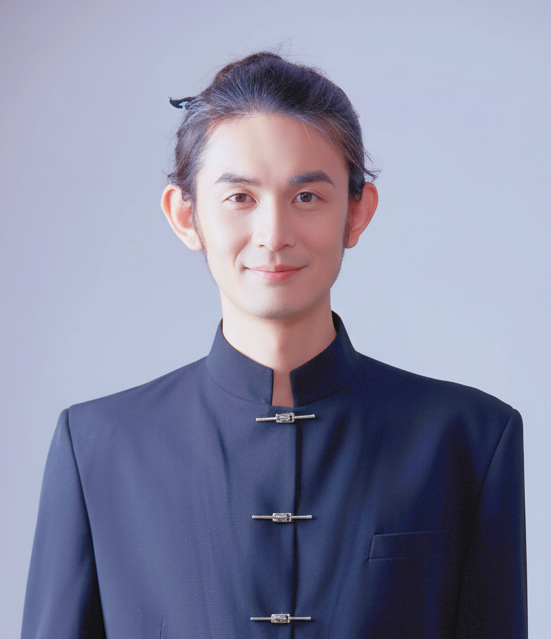}}]{Yongqi Dong}
received the M.S. degree in control science and engineering from
Tsinghua University, Beijing, China, in 2017, and the Ph.D. degree in transport and planning from Delft University of Technology, Delft, the Netherlands, in
2025. He is currently working as a Research Group Leader at RWTH Aachen University, Aachen, Germany. He serves as the Chair for the IEEE ITSS Technical Committee on \textit{Automated Mobility in Mixed Traffic}. His research interests include deep learning, automated driving, vision-language-models, and traffic safety. He seeks to employ AI and interdisciplinary research as tools to shape a better world.

\end{IEEEbiography}

\end{document}